\pdfoutput=1

\documentclass[11pt]{article}

\usepackage[final]{acl}
\usepackage{colortbl}
\usepackage{import}
\usepackage{times}
\usepackage{markdown}
\usepackage{latexsym}
\usepackage[T1]{fontenc}
\usepackage[frozencache,cachedir=./cache]{minted}
\definecolor{promptColor1}{RGB}{245,245,220} 
\definecolor{promptColor2}{RGB}{220,235,255} 
\definecolor{promptColor3}{RGB}{232,245,219} 
\definecolor{promptColor4}{RGB}{255,243,224} 
\definecolor{promptColor5}{RGB}{253,236,234} 

\usepackage[utf8]{inputenc}
\usepackage{subfigure}
\usepackage{subcaption}
\usepackage{microtype}
\usepackage{booktabs}
\usepackage{inconsolata}
\usepackage{xcolor}
\usepackage{xspace}
\usepackage{comment}
\usepackage{color, colortbl}

\newcommand{\lang}[1]{\texttt{#1}}
\newcommand{\dataset}[0]{\textsc{Injongo}}
\newcommand*{\clinc}{\textsc{Clinc} \xspace}
\newcommand*{\massive}{\textsc{Massive} \xspace}

\newcommand*{\gemini}{Gemini 1.5 Pro \xspace}
\newcommand*{\gpto}{GPT-4o\xspace}
\newcommand*{\yoruba}{Yor\`ub\'a\xspace}
\usepackage{graphicx}
\usepackage{multirow}
\usepackage{float}
\definecolor{Gray}{gray}{0.9}
\definecolor{LightCyan}{rgb}{0.88,1,1}
\usepackage{listings}
\title{\dataset: A Multicultural Intent Detection and Slot-filling Dataset for 16 African Languages}
\author{
\normalsize 
Hao Yu$^{1,2}$, Jesujoba O. Alabi$^{3,*}$, Andiswa Bukula$^{4,*}$, Jian Yun Zhuang$^{5}$, En-Shiun Annie Lee$^{6}$,\\ 
\textbf{\normalsize Tadesse Kebede Guge$^{*}$, Israel Abebe Azime$^{3}$, Happy Buzaaba$^{7,*}$, Blessing Kudzaishe Sibanda$^{*}$},\\ 
\textbf{\normalsize Godson K. Kalipe$^{*}$, Jonathan Mukiibi$^{8,*}$, Salomon Kabongo Kabenamualu$^{9,*}$, Mmasibidi Setaka$^{4}$},\\ 
\textbf{\normalsize Lolwethu Ndolela$^{*}$, Nkiruka Odu$^{*}$, Rooweither Mabuya$^{4,*}$, Shamsuddeen Hassan Muhammad$^{10,*}$},\\ 
\textbf{\normalsize Salomey Osei$^{11}$, Sokhar Samb$^{12}$, Juliet W. Murage$^{*}$, Dietrich Klakow$^{3}$, David Ifeoluwa Adelani$^{1,2,*}$} \\ 
\textbf{\normalsize } \\
\footnotesize
$^*$Masakhane NLP, 
$^{1}$McGill University, Canada, 
$^{2}$Mila, Quebec AI Institute, Canada, 
$^{3}$Saarland University, Germany,\\
\footnotesize
$^{4}$SADiLaR, South Africa, 
$^{5}$University of Toronto, Canada, 
$^{6}$ OntarioTech University, Canada,
$^{7}$Princeton University, USA,\\
\footnotesize
$^{8}$Makerere University, Uganda, 
$^{9}$L3S Research Center, Germany, 
$^{10}$Imperial College London, United Kingdom,\\
\footnotesize
$^{11}$Universidad de Deusto, Spain,
$^{12}$Dakar American University Of Science and Technology, Senegal.
\vspace{5em}
}
\begin{document}
\maketitle

\begin{abstract}
Slot-filling and intent detection are well-established tasks in Conversational AI. However, current large-scale benchmarks for these tasks often exclude evaluations of low-resource languages and rely on translations from English benchmarks, thereby predominantly reflecting Western-centric concepts. In this paper, we introduce \textsc{Injongo}---a multicultural, open-source benchmark dataset for 16 African languages with utterances generated by native speakers across diverse domains, including banking, travel, home, and dining. Through extensive experiments, we benchmark the fine-tuning multilingual transformer models and the prompting large language models (LLMs), and show the advantage of leveraging African-cultural utterances over Western-centric utterances for improving cross-lingual transfer from the English language. Experimental results reveal that current LLMs struggle with the slot-filling task, with GPT-4o achieving an average performance of 26 F1-score. In contrast, intent detection performance is notably better, with an average accuracy of 70.6\%, though it still falls behind the fine-tuning baselines. When compared to the English language, GPT-4o and fine-tuning baselines perform similarly on intent detection, achieving an accuracy of approximately 81\%. Our findings suggest that the performance of LLMs is still behind for many low-resource African languages, and more work is needed to further improve their downstream performance. 

\end{abstract}

\section{Introduction}

Intent detection and slot-filling are crucial components of the natural language understanding module in task-oriented dialogue systems~\citep{hemphill-etal-1990-atis, Coucke2018SnipsVP,gupta-etal-2018-semantic-parsing}. They map a user's request to a predefined semantic category recognized by the dialogue manager, along with extracting specific entities (known as slots). This process facilitates generating an appropriate response for the end user. Despite their importance, only a few languages have labeled datasets available for these tasks across multiple domains~\citep{Larson2022ASO}. 

Several efforts have been made to make datasets multilingual through human translation into other languages~\citep{xu-etal-2020-end,li-etal-2021-mtop,van-der-goot-etal-2021-masked,ruder-etal-2023-xtreme}. However, these efforts face two key challenges: (1) the translationese effect, which makes utterances sound less natural in the target languages~\citep{vanmassenhove-etal-2021-machine,bizzoni-etal-2020-human}, and (2) the creation of utterances that are less culturally relevant. The Massive dataset \cite{fitzgerald-etal-2023-massive}, which covers 51 languages, addresses the second challenge by encouraging translators to ``localize'', ``translate'', or ``keep the slot unchanged''. Despite improvements in the utterance generation process, \textsc{Massive} includes only three African languages (Amharic, Afrikaans and Swahili), and many utterances remain culturally irrelevant to the target language communities.

In this paper, we develop \dataset{}---the first large-scale \textit{multicultural} intent
detection and slot-filling dataset covering 16 African languages, and English language. We cover the following five domains: banking, home, travel, utility, and kitchen \& dining. The data construction process starts with providing an annotator with sentences from the \clinc dataset~\citep{Larson2022ASO} with a specified \textit{intent type}, and they are to come up with culturally-relevant similar sentences and relevant slot entities (see \autoref{fig:example}). The utterance generation process is followed by slots annotation. \dataset{} dataset covers 5 domains, 40 intents, 23 slots, and 3,200 instances per African language. 

\begin{table*}[t]
\small\centering
\resizebox{\textwidth}{!}{%
  \begin{tabular}{lrrrrrll}
  \toprule
  \textbf{Dataset} & \textbf{\# Domains} & \textbf{\# Intents}  & \textbf{\# Slots} & \textbf{\# utterances} & \textbf{\# Languages} & \textbf{\# African languages} & \textbf{Multi-cultural?} \\
  \midrule
CLINC~\citep{CLINC}& 10 & 150 & 0 & 23,700 & 1 & 0 & yes \\
Facebook~\citep{schuster-etal-2019-cross-lingual} & 3 & 12 & 11 & 57,000 & 3 & 0 & yes \\
MultiATIS~\citep{xu-etal-2020-end} & 11 & 26 & 140 & 44,943 & 9 & 0 & no \\
xSID~\cite{van-der-goot-etal-2021-masked} & 7 & 16 & 33 & 10,000 & 13 & 0 & no \\
MTOP~\cite{li-etal-2021-mtop} & 11 & 117 & 78 & 100,000 & 6 &  0 & no \\
MTOP++~\citep{ruder-etal-2023-xtreme} & 11 & 117 & 78 & 144,243 & 20 & 5 (amh, hau, yor, swa, zul) & no \\
MASSIVE~\citep{fitzgerald-etal-2023-massive} & 18 & 60 & 55 & 995,571 & 51 & 3 (afr, amh, swa) & partial\\
\midrule
\dataset~(Ours) & 5 & 40 & 23 & 52,979 & 17 & \textbf{16} & yes \\
  \bottomrule
  \end{tabular}
}
  \vspace{-2mm}
  \caption{\textbf{Overview of important related works that intent detection and slot-filling tasks}. We included the number of domains, intents, slots, languages, African languages and how multicultural are the utterances. }
  \label{tab:past_works}
\end{table*}

We performed several supervised fine-tuning experiments with multilingual encoders and prompting of Large Language Models (LLMs), both using \dataset{}. Our result shows that fine-tuning baselines could reach an accuracy of 93.7\% and F1-score of 85.6 for intent detection and slot-filling tasks respectively. While the best prompting of LLMs results (GPT-4o) drops by -28\% accuracy point and $-52.6$ F1 score. While slot-filling and named entity recognition tasks are often challenging for LLMs even for English~\citep{Yu2023OpenCO}, intent detection performance in English is similar performance whether we use fine-tuning baselines or prompt GPT-4o. Our findings suggest that LLMs performance is still behind for many low-resource African languages, and more work is needed to further improve their downstream performance.  For reproducibility, we open-source our code\footnote{\href{https://github.com/McGill-NLP/Injongo}{McGill-NLP/Injongo}} and dataset \footnote{\href{https://github.com/masakhane-io/masakhane-nlu/tree/main/InjongoIntent}{Masakhane-NLU}} on GitHub. Dataset is released under CC BY 4.0 license. Models will be released on the HuggingFace soon. 
 


\section{Related Work}

\paragraph{African Benchmarks}
Limited available labeled datasets are one of the major challenges of AfricaNLP. Since 2021, there have been many grassroots efforts to create large-scale datasets for African languages covering several tasks such as machine translation~\citep{Alabi2025}, named entity recognition~\citep{adelani-etal-2021-masakhaner,adelani-etal-2022-masakhaner}, sentiment classification~\citep{muhammad-etal-2023-afrisenti}, hate speech~\citep{Muhammad2025AfriHateAM}, question answering~\citep{ogundepo-etal-2023-cross}, topic classification~\citep{adelani-etal-2023-masakhanews,AfroXLM-76L} covering 10 to 57 languages. 
The closest benchmark to our task of slot-filling is the MasakhaNER~\citep{adelani-etal-2021-masakhaner,adelani-etal-2022-masakhaner} that covers 20 African languages but they focus on four entity types  ``personal names'', ``organization'', ``location'', and ``dates'', which are not fine-grained and well adapted to several domains such as banking and travel 
that we cover in \dataset{}.



\paragraph{Intent and Slot-filling Benchmarks} Most of the existing benchmarks for intent detection and slot-filling tasks are English-only. There are a few efforts to make them multilingual in two ways: (1) human generating the utterances in a particular domain, followed by intent and slot filling annotation. (2) through human translation of annotated data from English to other languages which introduces some cultural bias since Western entities are being propagated. While the first approach is the most ideal methodology, it is very cost-intensive when scaling to many languages. Facebook dataset~\citep{schuster-etal-2019-cross-lingual} followed the first approach by creating labeled data in three domains (alarm, reminder and weather) for three languages: English, Spanish and Thai. However, most other approaches make use of the second approach, where English data are translated to other languages~\citep{xu-etal-2020-end, van-der-goot-etal-2021-masked,li-etal-2021-mtop}, however, they often do not include African languages. XTREME-UP benchmark expanded the MTOP dataset~\citep{li-etal-2021-mtop} to five African languages (Amharic, Hausa, Yoruba, Swahili and Zulu), while \massive~\citep{fitzgerald-etal-2023-massive} perform human translation to 50 languages including three African languages (Afrikaans, Amharic, and Swahili). \massive benchmark partially addresses this Western cultural bias by encouraging translators to replace entities with more culturally relevant ones, but Western entities are still prevalent in the dataset. \autoref{tab:past_works} summarizes all existing related works. In our paper, we introduce \dataset{} which is the largest intent detection and slot-filling dataset covering 16 African languages, and we ensured that the slot entities are more culturally relevant in the respective countries the languages are from.

\section{Introducing \dataset{} Dataset}

\begin{table}[t]
  \centering
  \resizebox{\columnwidth}{!}{%
    \begin{tabular}{lclr}
    \toprule[1pt]
    \textbf{Language} & \textbf{Code} & \textbf{Language Family} & \textbf{No. of Speakers} \\ \midrule
    Amharic & amh & Afro-Asiatic/Semitic & 60M \\
    Ewe & ewe & Niger-Congo/Kwa & 7M \\
    Hausa & hau & Afro-Asiatic/Chadic & 63M \\
    Igbo & ibo & Niger-Congo/Volta-Niger & 27M \\
    Kinyarwanda & kin & Niger-Congo/Bantu & 10M \\
    Lingala & lin & Niger-Congo/Bantu & 41M \\
    Luganda & lug & Niger-Congo/Bantu & 7M \\
    Oromo & orm & Afro-Asiatic/Cushitic & 46M \\
    Shona & sna & Niger-Congo/Bantu & 12M \\
    Sesotho & sot & Niger-Congo/Bantu & 7M \\
    Swahili & swa & Niger-Congo/Bantu & 98M \\
    Twi & twi & Niger-Congo/Kwa & 9M \\
    Wolof & wol & Niger-Congo/Senegambia & 5M \\
    Xhosa & xho & Niger-Congo/Bantu & 9M \\
    Yoruba & yor & Niger-Congo/Volta-Niger & 42M \\
    Zulu & zul & Niger-Congo/Bantu & 27M \\
    \bottomrule[1pt]
    \end{tabular}%
  }
  \vspace{-2mm}
  \caption{\textbf{Overview of languages in the \dataset{} dataset}, including ISO 639-3 language codes, language families, and approximate number of speakers.}
  \label{tab:lang-overview}
\end{table}

\begin{figure*}[!ht]
    \centering
    \includegraphics[width=1\linewidth]{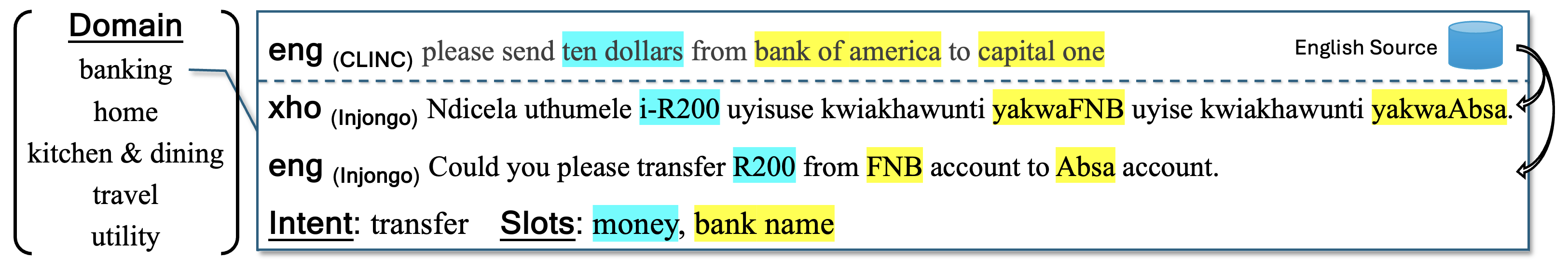}
    \vspace{-5mm}
    \caption{\textbf{Task description for \dataset{} dataset}. An example from one of the five domains. It shows the semantic-similar sentences along with intent and slot-filling labels.}
    
    \label{fig:example}
\end{figure*}

\dataset{}\footnote{\dataset{} means intent in isiXhosa language.} is a joint intent detection and slot-filling dataset (\textbf{ID-SF}) for typologically diverse Sub-Saharan African languages and English. The selected languages represent diverse linguistic families and are widely spoken in Africa. These languages come from the two dominant language families in Africa: 13 from Niger-Congo and three from Afro-Asiatic. The languages covered are spoken by a large population in Africa, ranging from Swahili with 98M speakers to Wolof with 5M speakers, making the dataset particularly valuable for over 400 million African population. ~\autoref{tab:lang-overview} shows the languages covered, their language family, and the number of speakers of the languages.

\subsection{Data source and collection}
Typical ID-SF data collection often requires large crowd-sourcing efforts to collect utterances, with additional labeling of intents and slots in various domains. Developing such a large crowd-sourcing effort is time-consuming and costly for several low-resource languages. To simplify the process while making the dataset cultural, we provide each annotator with sample sentences from the \clinc dataset~\citep{Larson2022ASO} with a specified \textit{intent type}, say ``transfer''. Then, the dataset construction follows two stages: (1) \textbf{Utterance elicitation} in an African language and (2) \textbf{Slot-filling annotation} of the generated utterance. 

~\autoref{fig:example} shows an example of an English utterance from the \clinc dataset in the banking domain: ``\textit{please send ten dollars from bank of america to capital one}''. The corresponding intent label is ``\textit{transfer}'', and the entities of slot filling are the amount of [money] (\textit{ten dollars}), the source [bank] (\textit{bank of america}), and the destination [bank] (\textit{capital one}). A Xhosa annotator was asked to generate another utterance belonging to the same intent type but capturing the South African context where the language is spoken. Thus, the annotator used the R200 as \textit{``money''} with currency Rand (R), and more familiar South African banks such as ``FNB'' and ``Absa'' for \textit{``bank name''} slot. We provide more information about the two stages of data construction below.

\paragraph{Utterance generation}
The source data for our multilingual benchmark is from the \clinc English dataset---an intent detection with 150 intent classes across 10 domains (but without slot annotation)~\footnote{The domains are: banking, work, meta, auto \& commute, travel, home, utility, kitchen \& dining, small talk, and credit cards}, we extracted 40 intents from five most suitable domains to the African contexts: \textbf{Banking} (e.g. ``transfer'',  ``pay bill''),  \textbf{Home} (e.g. ``play music'', ``calendar update''),  \textbf{Kitchen and Dining} (e.g. ``recipe'', ``confirm reservation''), \textbf{Travel} (e.g. ``exchange rate'', ``book flight'' ), and \textbf{Utility} (e.g. ``alarm'', ``make call'' ). Next, we conducted the tutorial on the utterance generation task and a \textbf{practice session} and asked every annotator to generate a sample English utterance per intent that culturally aligns with the African contexts (e.g. food type or language name). Per language, we recruited three annotators, and they generated 120 utterances (40 per annotator and intent). We aggregated the practice data as the \textit{\dataset{} English dataset}. 
Finally, for the \textbf{full data collection}, we asked the same three annotators to generate \textit{80 utterances per intent}, given a sample sentence from \clinc. Each annotator worked on different intents. In total, we collected 3,200 utterances with a balanced number of intent types. Appendix~\ref{sec:intent-label} contains all the 40 intent types selected.

\begin{table}[t]
  \centering
  \resizebox{0.48\textwidth}{!}{%
    \begin{tabular}{l|ccccc}
    \toprule[1pt]
    Lang. & Total & Avg. & Un. Fleiss'$\kappa$ & Fleiss'$\kappa$ & $\Delta$ \\ \midrule
    \lang{amh} & 10555 & 3.30 & 0.850 & 0.935 & +0.085 \\
    \lang{ewe} & 11181 & 3.49 & 0.875 & 1.000 & +0.125 \\
    \lang{hau} & 11491 & 3.59 & 0.892 & 0.997 & +0.105 \\
    \lang{ibo} & 12246 & 3.82 & 0.812 & 0.973 & +0.161 \\
    \lang{kin} & 10112 & 3.16 & 0.740 & 0.963 & +0.224 \\
    \lang{lin} & 11025 & 3.44 & 0.823 & 0.990 & +0.168 \\
    \lang{lug} & 11769 & 3.67 & 0.888 & 0.990 & +0.102 \\
    \lang{orm} & 11958 & 3.74 & 0.849 & 0.992 & +0.143 \\
    \lang{sna} & 15222 & 4.76 & 0.935 & 0.976 & +0.041 \\
    \lang{sot} & 6468 & 2.02 & 0.694 & 0.997 & +0.303 \\
    \lang{swa} & 14217 & 4.44 & 0.878 & 0.986 & +0.107 \\
    \lang{twi} & 14325 & 4.48 & 0.916 & 0.986 & +0.070 \\
    \lang{wol} & 10942 & 3.42 & 0.728 & 0.942 & +0.213 \\
    \lang{xho} & 12475 & 3.90 & 0.825 & 0.938 & +0.113 \\
    \lang{yor} & 13620 & 4.26 & 0.862 & 0.988 & +0.126 \\
    \lang{zul} & 11911 & 3.73 & 0.640 & 0.913 & +0.273 \\ 
    \bottomrule[1pt]
  \end{tabular}%
  }
  \vspace{-2mm}
  \caption{\textbf{Statistics of slot entity annotations across languages}. For each language, we show the total number of annotated entities, average entities per sentence, and inter-annotator agreement measured by Fleiss' Kappa ($\kappa$) before (Un.) and after review. $\Delta$ shows the improvement in agreement after the review process.}
  \label{tab:dataset-fk-merged}
\end{table}

\paragraph{Slot-filling annotation}
Similar to the utterance generation phase, we first conducted a practice session in English to train annotators followed by the full data annotation. We manually analyzed each generated utterance to come up with the most relevant slot entities (about 26). However, after the practice session, annotators recommended the addition of new slots such as ``airline'', ``airport name'', ``car type'', and ``supermarket name'', which we adopted. After the practice session, we gave detailed feedback on the issues with the annotation, and annotators discussed with their language coordinator how to resolve issues. Finally, we asked them to annotate the slot entities for the 3,200 utterances. Each utterance was annotated by three annotators so that we could check for agreement in the slot annotations. The annotation followed the named entity recognition setup on LabelStudio platform \footnote{\url{https://labelstud.io/}}. Appendix ~\ref{sec:slot-label} contains all the 34 intent types selected. 

For both utterance elicitation and slot-filling annotation, all recruited participants received an appropriate remuneration based on the per-country rate decided by our logistic company in Kenya.\footnote{Utterance elicitation rate ranges from \$1,555 to \$2,838 per language depending on country rate, and slot-filling annotation ranges from \$388 to \$709}




\begin{figure*}[h]
    \centering
    \includegraphics[width=0.95\linewidth]{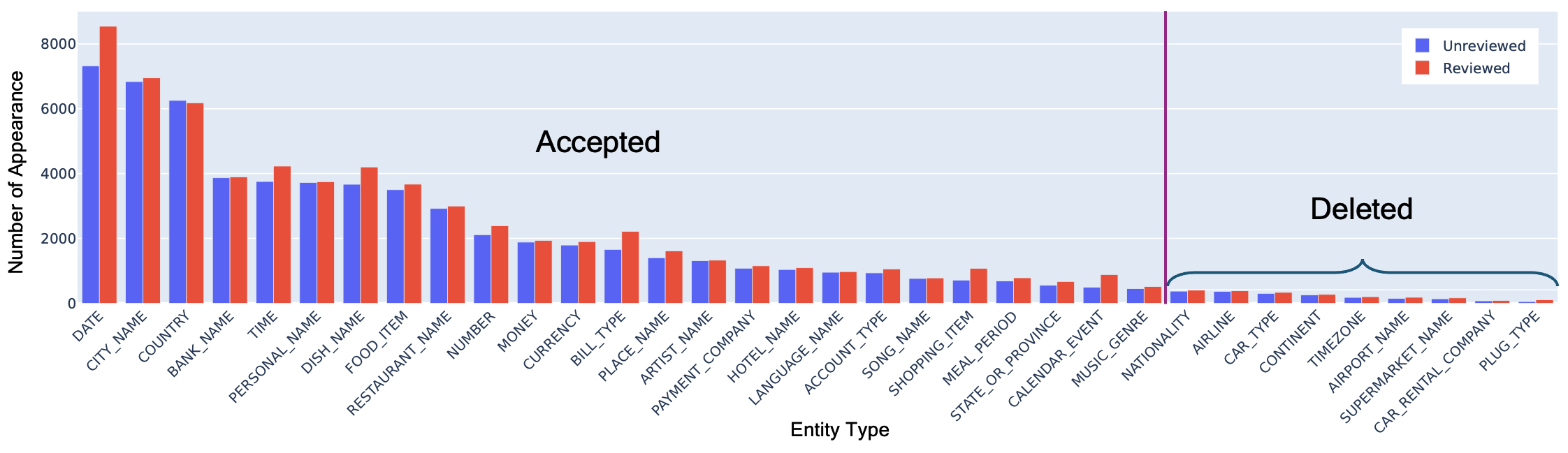}
    \vspace{-3mm}
    \caption{\textbf{The distribution of slot entities appearances of all 16 African languages} with Unreviewed and Reviewed versions. The slot entities are sorted from left to right by frequency in descending order.}
    \label{fig:entity-dist}
\end{figure*}

\subsection{Quality Control for Slot-filling}
\label{sec:quality-control}
To ensure annotation quality and consistency, we follow a rigorous quality control process using a majority voting system with a minimum of three annotators per sentence to resolve disagreements. The annotation quality was evaluated using Fleiss' Kappa score \citep{Fleiss}, with scores presented in \autoref{tab:dataset-fk-merged} comparing agreement levels before and after the review process. 
Initial Fleiss' Kappa scores revealed substantial variation across languages, ranging from 0.618 (Zulu) to 0.934 (Shona), indicating significant inter-annotator disagreement. Following the review process, agreement scores improved markedly across all languages, reaching 0.912-1.00. Notable improvements were observed in Sesotho (+0.327) and Zulu (+0.294), with other languages showing average improvements of approximately 0.1 in their Fless' Kappa scores.

\subsection{Slot-filling label merging}
\label{sec:label-merge}
On completion of the final annotation, we found that some slot entities are rarely used. We performed an analysis of entity frequency distribution across all languages. \autoref{fig:entity-dist} shows the result of our analysis, we decided to exclude slot entities appearing less than 500 times across all languages (after \textit{MUSIC GENRE} in the figure).  Consequently, nine infrequent slots from \textit{NATIONALITY} through \textit{PLUG TYPE} were eliminated. Examination of annotator feedback and comparative analysis between unreviewed and reviewed versions indicated that ambiguous slot types significantly impacted annotation quality and introduced unnecessary complexity. To enhance annotation clarity and maintain consistency, the following merging strategy was implemented:
\begin{itemize}{\leftmargin=0.5mm}\setlength{\itemsep}{0em}\setlength{\parskip}{0em}
    \item \textit{Geographic entities}: \textit{STATE OR PROVINCE} and \textit{CITY NAME} were consolidated into a unified \textit{CITY OR PROVINCE} category to ensure consistent handling of geographic references.
    \item \textit{Food-Related Labels}: \textit{DISH NAME} and \textit{FOOD ITEM} were unified under \textit{DISH OR FOOD} to eliminate classification ambiguity.
\end{itemize}
This merging process resulted in a reduction from 34 to 23 slot types. The complete enumeration of original and consolidated labels, along with unmerged entity Fleiss' kappa scores, is provided in Appendix \ref{app:stat}.

\begin{table}[t]
  \centering
  \footnotesize
  \resizebox{\columnwidth}{!}{%
    \begin{tabular}{lcc|c}
    \toprule[1pt]
     & \multicolumn{2}{c}{\textbf{\dataset{}}} & \multicolumn{1}{c}{\textbf{\clinc}}  \\    
    \textbf{split} & \textbf{African} & \textbf{English} & \textbf{English} \\
    \midrule
    TRAIN & 2,240 (56 per intent)  & 1,047 & 4,000 (100 per intent) \\
    DEV & 320 (8 per intent) & 110 & 800 (20 per intent) \\
    TEST & 640 (16 per intent) & 622 & 1,200 (30 per intent) \\
    \bottomrule
    \end{tabular}%
  }
  \vspace{-2mm}
  \caption{\textbf{\dataset{} dataset split}. The African data have an equal number of samples per intent while the English samples per intent vary. }
  \label{tab:data_split}
\end{table}

\subsection{Data split}
Our final annotation resulted in 3,200 annotated utterances, with 80 utterances per intent for each of the 16 African languages. The dataset is partitioned following ratios of 70\%, 10\%, and 20\% for train, dev, and test splits respectively, stratified by intent for each language. Additionally, we aggregated the practice utterances generated and the practice slot annotations as the English dataset, leading to 17 annotated languages. In total, the English consist of 1779 utterances.~\footnote{Ideally, if each language completes 120 utterance generation, we ought to have 1920 utterances, however, some languages only did 80 in the practice, leading to a slightly lower English portion.} Finally, we sampled 4000 \clinc intent-only dataset to compare western-centric English dataset to our curated \dataset{} dataset that captures the African contexts. \autoref{tab:data_split} provides the comprehensive dataset statistics of the African languages and English splits.

\section{Experiments Setup}

\subsection{Fine-tuning baselines}
We evaluate three categories of models: (1) \textbf{encoder-only models} such as XLM-RoBERTa Large \cite{XLM-R}, AfroXLMR~\cite{AfroXLM}, AfroXLMR-76L~\cite{AfroXLM-76L}, AfriBERTa V2~\cite{oladipo2024scaling}, (2)  \textbf{encoder-decoder models} such as mT5-Large \cite{mT5}, AfriTeVa V2 Large \cite{AfriTeVaV2}, and (3)  \textbf{a multilingual variant of LLM2Vec} model~\citep{llm2vec} i.e. NLLB-LLM2Vec \cite{NLLB-LLM2Vec} that stack NLLB-encoder~\citep{NLLB} model with LLaMa 3.1 8B~\citep{Llama3} to develop a multilingual sentence transformer model. These models are fine-tuned using the AdamW optimizer for 20 epochs with early stopping. All results are averaged over five runs. Learning rates are calibrated for each architecture and task as detailed in Appendix \ref{sec:lr-impact}. The languages covered in each pre-trained model are available in Appendix \ref{app:language}. 

\begin{table*}[!ht]
\centering
\resizebox{\textwidth}{!}{%
\begin{tabular}{c|l|cccccccccccccccccc}
\toprule[1pt]
\multicolumn{1}{l|}{\textbf{Task}} & \textbf{Model} & \lang{eng} & \lang{amh} & \lang{ewe} & \lang{hau} & \lang{ibo} & \lang{kin} & \lang{lin} & \lang{lug} & \lang{orm} & \lang{sna} & \lang{sot} & \lang{swa} & \lang{twi} & \lang{wol} & \lang{xho} & \lang{yor} & \lang{zul} & \textbf{AVG} \\ \midrule
& \multicolumn{19}{l}{\textbf{\texttt{In-language training}}} \\ 
\multirow{10}{*}{\begin{tabular}[c]{@{}c@{}} \textsc{Intent} \\ \textsc{Detection}\end{tabular}} & mT5-Large & 80.5 & 91.5 & 77.3 & 94.6 & 92.9 & 83.7 & 91.3 & 83.3 & 73.3 & 92.6 & 80.2 & 95.8 & 85.3 & 91.6 & 95.8 & 90.9 & 82.4 & 87.7$_{\pm 4.1}$ \\
& AfriTeVa V2 (T5) & 81.6 & 93.2& 84.4& \textbf{98.9}& 95.7& 87.8& 91.6& 86.8& 86.6& 94.6& 85.7& 96.8& 87.1& 94.0& 97.3& 97.0& 89.2& 91.7$_{\pm 2.7}$ \\
\cmidrule{2-20}
& NLLB LLM2Vec& \textbf{88.4} & 94.2& 87.8& 98.3& \textbf{96.8}& 89.2& \textbf{95.2}& \textbf{93.2}& 86.2& \textbf{96.1}& 87.3& 97.4& 93.5& 95.6& \textbf{97.5}& 97.3& 89.1& 93.4$_{\pm 2.3}$ \\
& XLM-RoBERTa& 83.5 & 92.9& 77.9& 96.0& 88.8& 69.6& 90.5& 78.9& 75.0& 83.8& 76.0& 96.7& 79.5& 90.2& 89.6& 92.6& 74.7& 84.5$_{\pm 4.9}$ \\
& AfriBERTa V2& 74.2 & 91.2& 78.3& 98.2& 93.8& 83.1& 91.0& 83.8& 78.8& 89.5& 81.9& 96.0& 83.2& 92.3& 94.4& 95.0& 86.7& 88.6$_{\pm 3.5}$ \\
& AfroXLMR& 84.1 & 95.3& 84.6& 98.3& 96.0& 88.2& 93.3& 85.2& 88.3& 95.3& 85.5& 97.8& 88.8& 95.8& 97.3& 96.1& 89.0& 92.2$_{\pm 3.0}$ \\
\rowcolor{Gray}
\cellcolor{white} & AfroXLMR 76L& 84.5 & \textbf{95.5}& \textbf{90.4}& 98.7& 96.3& \textbf{89.4}& 94.6& 91.3& \textbf{88.3}& 95.1& \textbf{86.8}& \textbf{98.1}& \textbf{93.6}& \textbf{96.2}& 96.9& \textbf{97.7}& \textbf{89.8}& \textbf{93.7}$_{\pm 2.1}$ \\
\cmidrule{2-20}
& \multicolumn{19}{l}{\textbf{\texttt{Multi-lingual training}}} \\ \rowcolor{LightCyan}
\cellcolor{white} & \multicolumn{1}{l|}{AfroXLMR-76L}& 89.0 & 96.0& 92.6& 99.2& 96.6& 87.7& 95.9& 92.3& 92.9& 96.5& 87.6& 97.8& 94.2& 97.1& 97.3& 97.9& 89.2& \textbf{94.4$_{\pm 2.0}$} \\ \midrule
& \multicolumn{19}{l}{\textbf{\texttt{In-language training}}} \\ 
\multirow{8}{*}{{\begin{tabular}[c]{@{}c@{}}\textsc{Slot} \\\textsc{Filling}\end{tabular}}} & mT5-Large& 73.7 & 80.9& 71.6& 89.4& 80.5& 74.2& 82.6& 78.9& 72.1& 81.1& 74.7& 88.1& 79.0& 76.9& 88.4& 78.9& 68.3& 79.1$_{\pm 3.7}$ \\
& AfriTeVa V2 (T5) & 73.6 & 80.9& 74.5& 93.8& 79.9& 76.6& 87.1& 85.2& 79.0& 82.1& \textbf{77.5}& 88.9& 84.0& 79.0& 90.0& 87.2& 71.2& 82.3$_{\pm 3.3}$ \\
\cmidrule{2-20}
& NLLB LLM2Vec& 74.6 & 82.4& 80.5& 93.6& 78.1& 70.1& 84.8& 86.6& 80.8& 81.4& 74.8& 85.7& 85.7& 78.3& 88.0& 85.0& 78.3& 82.1$_{\pm 3.1}$ \\
& XLM-RoBERTa & 77.9 & 84.8& 79.9& 93.9& 76.6& 69.3& 86.3& 83.8& 83.8& 79.3& 71.7& 88.7& 84.2& 79.3& 89.1& 83.9& 79.4& 82.1$_{\pm 3.5}$ \\
& AfriBERTa V2 & 70.7 & 82.2& 77.9& 93.7& 78.3& 73.8& 84.4& 84.1& 81.0& 81.8& 73.5& 87.6& 81.9& 78.3& 88.5& 86.2& 79.6& 82.1$_{\pm 2.9}$ \\
& AfroXLMR& \textbf{79.0} & 86.2& 81.6& \textbf{95.1}& \textbf{82.0}& 76.3& 87.1& 88.5& 84.9& \textbf{84.9}& \textbf{77.5}& \textbf{90.2}& 85.5& \textbf{81.7}& \textbf{91.1}& 87.3& \textbf{82.5}& 85.2$_{\pm 2.7}$ \\
\rowcolor{Gray}
\cellcolor{white} & AfroXLMR 76L& 78.7 & \textbf{86.3}& \textbf{84.5}& 94.3& 81.9& \textbf{76.7}& \textbf{88.0}& \textbf{88.8}& \textbf{85.5}& \textbf{84.9}& 77.4& \textbf{90.2}& \textbf{89.8}& 81.3& 90.5& \textbf{88.1}& 81.3& \textbf{85.6$_{\pm 2.7}$} \\
\cmidrule{2-20}
& \multicolumn{19}{l}{\textbf{\texttt{Multi-lingual training}}} \\ 
\rowcolor{LightCyan}
\cellcolor{white} & \multicolumn{1}{l|}{AfroXLMR 76L}& 82.4 & 88.2& 87.0& 96.3& 84.0& 79.3& 90.3& 89.2& 87.2& 86.1& 80.4& 90.5& 90.3& 83.3& 91.8& 90.2& 83.3& \textbf{87.3$_{\pm 2.4}$} \\
\bottomrule[1pt]
\end{tabular}%
}
\vspace{-2mm}
\caption{\textbf{Intent detection and slot-filling results for supervised fine-tuned Small LMs on \dataset{}}. Models are ranked by \underline{accuracy} for \underline{intent detection} and \underline{F1-score} for \underline{slot-filling}. The average performance and standard deviation across 16 African languages are reported. The best models are highlighted in \textbf{Gray} and \textbf{Cyan} colours.}
\label{tab:sft-result}
\end{table*}


\subsection{LLM Prompting} 
First, we conduct \textbf{zero-shot prompting} using the following widely used LLMs for evaluation: GPT-4o,\footnote{\url{https://platform.openai.com/docs/models\#gpt-4o}} Gemini 1.5 Pro~\citep{Reid2024Gemini1U}
, Gemma 2 9B/27B Instruct \cite{Gemma2}, Llama 3.1 8B/3.3 70B Instruct~\cite{Llama3}, and Aya-101~\cite{aya}. We make use of five different prompts for each LLM. Second, we perform \textbf{few-shot evaluation} using the best-performing zero-shot template for each task (see Appendix \ref{app:prompts}). We employ two few-shot strategies (1) \textit{5-examples}: prompting with any 5 samples from different domains (see \autoref{fig:example}) i.e. one intent type is covered by domain (2) \textit{One-shot} intent-type prompting i.e. one sample per intent type or 40 samples from different intent types. We used the same samples for both tasks. Finally, we extend to 4 shots ---acceptable maximum context length (CL) for Gemma 2, Aya-101 was excluded for small CL.

Finally, as an additional strong baseline for LLMs, we performed supervised fine-tuning (SFT) on the Gemma 2 9B for 5 epochs using learning rates of $2\times10^{-5}$/ $2.5\times10^{-5}$ for intent detection and slot filling. The dataset of SFT was obtained by aggregating all the training samples of the 17 languages in \dataset{} i.e. ``Combined \dataset{}'', with randomly sampled prompts from a pool of 5. The evaluations of LLMs use 5 different prompting templates and a temperature of 0.5. We provide all the prompts used in Appendix \ref{app:prompts}.

\subsection{Cross-lingual Transfer Analysis}
To investigate how well our dataset facilitates cross-lingual learning and transfer capabilities across languages,
we tested two settings (1) \textbf{zero-shot transfer} from the English split of \dataset{}, and evaluated on African languages. (2) \textbf{Translate-Test} where we evaluate the best English model on the machine-translated sentence test sets from an African language to English. We leveraged the NLLB-200-3.3B~\cite{NLLB}  machine translation model for the Translate-test setting. We compare the results with LLM prompting.

\paragraph{Hyper-parameters and Prompts used} 
Experiments of the baselines and cross-lingual transfer runs make use of five fixed random seeds. 
Detailed experiments setup, training configuration and prompts are in Appendix \ref{sec:exp-more}.

\begin{table*}[!htpb]
\centering
\resizebox{\textwidth}{!}{%
\begin{tabular}{l|l|cccccccccccccccccc}
\toprule
\textbf{Task} & \textbf{Model} & \lang{eng} & \lang{amh} & \lang{ewe} & \lang{hau} & \lang{ibo} & \lang{kin} & \lang{lin} & \lang{lug} & \lang{orm} & \lang{sna} & \lang{sot} & \lang{swa} & \lang{twi} & \lang{wol} & \lang{xho} & \lang{yor} & \lang{zul}& \textbf{AVG} \\ \midrule
\multirow{8}{*}{\begin{tabular}[c]{@{}c@{}} \textsc{Intent} \\ \textsc{Detection}\end{tabular}} 
& Llama 3.1 8B  & 27.6 & 1.9 & 2.1 & 4.8 & 5.5 & 3.3 & 5.3 & 2.4 & 1.6 & 2.8 & 2.9 & 14.1 & 2.6 & 4.0 & 3.2 & 3.5 & 2.8 & 3.9$_{\pm2.4}$ \\
& Gemma 2 9B & 77.6 & 49.2 & 6.1 & 40.8 & 31.5 & 23.8 & 22.2 & 23.2 & 7.7 & 29.7 & 19.9 & 70.0 & 21.0 & 13.8 & 40.1 & 32.2 & 36.3 & 29.2$_{\pm8.7}$ \\ 
& Aya-101 13B & 65.3 & 62.9 & 13.4 & 57.8 & 56.9 & 40.4 & 27.8 & 33.9 & 20.8 & 51.2 & 43.9 & 65.9 & 27.2 & 19.7 & 58.1 & 45.9 & 53.2 & 42.4$_{\pm9.1}$ \\
& Gemma 2 27B & 79.5 & 47.2 & 6.3 & 46.5 & 36.9 & 26.7 & 27.5 & 26.1 & 5.8 & 36.7 & 25.6 & 75.5 & 21.2 & 16.4 & 50.2 & 34.8 & 44.3 & 33.0$_{\pm9.6}$ \\
\rowcolor{Gray}
\cellcolor{white}
& Llama 3.3 70B  & 81.1 & 56.2 & 9.5 & 52.3 & 52.4 & 35.0 & 37.5 & 37.7 & 12.4 & 32.3 & 30.5 & 80.6 & 29.3 & 20.9 & 43.5 & 41.4 & 43.9 & 38.5$_{\pm9.5}$ \\
\cmidrule{2-20}
& Gemini 1.5 Pro & 81.8 & \textbf{77.9} & \textbf{24.3} & 74.8 & 65.4 & 61.5 & 54.6 & 59.3 & 39.3 & 68.6 & 51.6 & 83.2 & 47.2 & 25.6 & 76.2 & 66.8 & 68.7 & 59.1$_{\pm9.6}$ \\
\rowcolor{LightCyan}
\cellcolor{white}
& GPT-4o (Aug) & \textbf{80.9} & 76.0 & 15.1 & \textbf{80.7} & \textbf{71.8} & \textbf{64.7} & \textbf{56.4} & \textbf{68.2} & \textbf{59.3} & \textbf{75.5} & \textbf{59.7} & \textbf{84.5} & \textbf{58.6} & \textbf{43.7} & \textbf{79.6} & \textbf{77.0} & \textbf{71.2} & \textbf{65.1$_{\pm9.3}$} \\
\midrule
\multirow{8}{*}{\begin{tabular}[c]{@{}c@{}} \textsc{Slot} \\ \textsc{Filling}\end{tabular}} 
& Llama 3.1 8B & 25.0 & 3.7 & 5.6 & 11.1 & 12.6 & 8.5 & 9.1 & 10.1 & 2.8 & 9.9 & 11.5 & 17.3 & 11.2 & 9.2 & 2.6 & 11.0 & 9.0 & 9.1$_{\pm2.2}$ \\
& Gemma 2 IT 9B & 34.1 & 4.5 & 0.3 & 7.4 & 10.6 & 5.0 & 6.0 & 5.6 & 0.1 & 7.3 & 10.8 & 21.2 & 2.4 & 2.6 & 2.2 & 5.2 & 8.2 & 6.2$_{\pm2.9}$ \\ 
& Aya-101 13B & 21.4 & 8.2 & 7.9 & 11.8 & 14.6 & 12.2 & 9.4 & 15.5 & 3.6 & 15.0 & 17.0 & 16.2 & 13.8 & 14.0 & 2.8 & 9.6 & 10.6 & 11.4$_{\pm2.4}$ \\
& Gemma 2 IT 27B & 49.8 & 15.7 & 9.5 & 24.1 & 25.2 & 21.7 & 15.2 & 28.4 & 2.6 & 29.8 & 28.0 & 40.2 & 24.3 & 23.3 & 4.5 & 28.1 & 31.0 & 22.0$_{\pm5.8}$ \\
\rowcolor{Gray}
\cellcolor{white}
& Llama 3.3 70B Instruct & 52.6 & \textbf{26.3} & \textbf{22.0} & 29.5 & 35.0 & 31.4 & 25.0 & 30.4 & 9.3 & 29.5 & 36.4 & 40.7 & 35.6 & 36.4 & 6.9 & 34.2 & 31.9 & 28.8$_{\pm5.2}$ \\
\cmidrule{2-20}
& Gemini 1.5 Pro & 52.8 & 15.2 & 18.7 & 31.9 & 35.8 & 34.4 & \textbf{34.9} & 34.4 & 12.2 & 36.8 & \textbf{43.0} & 37.5 & 34.5 & 34.2 & 6.9 & 33.2 & \textbf{38.6} & 30.1$_{\pm6.1}$ \\
\rowcolor{LightCyan}
\cellcolor{white}
& GPT-4o (Aug) & \textbf{55.4} & 22.8 & 19.4 & \textbf{37.8} & \textbf{38.9} & \textbf{36.4} & 33.5 & \textbf{35.3} & \textbf{13.0} & \textbf{40.2} & 40.9 & \textbf{46.5} & \textbf{40.1} & \textbf{37.9} & \textbf{10.0} & \textbf{42.4} & 37.6 & \textbf{33.3$_{\pm6.0}$} \\

\bottomrule
\end{tabular}%
}
\vspace{-2mm}
\caption{\textbf{Zero-Shot performance of LLMs on Intent Detection (ID) and Slot Filling (SF)}. Evaluation is based on accuracy and F1-score for ID and SF tasks. Average computed on five templates, and on only African languages.}
\label{tab:prompt-results}
\end{table*}

\section{Results}

\subsection{Fine-tuned multilingual encoders}
\autoref{tab:sft-result} summarizes the results of the multilingual encoders fine-tuned \dataset{} dataset. Overall, AfroXLMR-76L achieves the best performance on both \textbf{ID-SF} tasks, with an average accuracy of 93.7\% and an F1 score of 85.6\%, respectively. We attribute the success of this model to the coverage of all languages in \dataset{} during its pre-training (see Appendix \autoref{tab:model_languages}). AfroXLMR, the earlier version of AfroXLMR-76L, follows closely with an average accuracy of 92.2\% and an F1 score of 85.2\%. However, this model was not pre-trained on some of the languages such as \texttt{ewe}, \texttt{twi}, \texttt{lin}, and \texttt{wol} leading to a significant drop in performance of $-5.8$, $-4.8$, $-1.3$, $-0.4$ for intent detection when compared to AfroXLMR-76L. This shows that multilingual encoders for African languages can significantly improve the performance over massively multilingual encoders covering more than 100 languages such as XLM-R and NLLB LLM2Vec. While NLLB LLM2Vec covers all languages in our dataset and is very effective for intent detection, it leads to $-3.5$ on slot-filling when compared to the performance of AfroXLMR-76L. In general, T5-based models such as mT5 and AfriTeVa V2 performed worse on both tasks compared to the BERT-based models, however, we still observe better performance of the African-centric T5-model, AfriTeVa V2 which gave decent results comparable to other models except AfroXLMR (-76L) models. 

Finally, we find that \textit{multilingual training} of AfroXLMR-76L over all languages gave better overall performance than \textit{in-language training} leading to $+0.7$ and $+1.7$ boost in performance on intent detection and slot-filling tasks respectively. This highlights the additional benefit of joint training of several languages, resulting in a \textit{single checkpoint} and better overall performance because they benefited from cross-lingual transfer learning among the languages. The languages that benefited the most are Oromo (\texttt{orm}) and English (\texttt{eng}) with $+4.6$ and $+4.5$ improvement respectively for intent detection. The large boost for English is because the training data is twice smaller than the remaining African languages ($1,047$ vs. $2,240$). Similarly, for slot-filling, the benefit of multilingual training is more obvious since all languages consistently improved in performance. We see that joint training benefited both high-resourced and low-resourced languages.

\subsection{LLMs prompting results}
\autoref{tab:prompt-results} shows the zero-shot LLM evaluation of five open models and two closed models. Our key findings are below: 

\paragraph{Slot-filling task is difficult for all LLMs including on English} The highest average performance achieved by the LLMs is $33.3$ for GPT-4o, although much better than the open model at $28.8$. We attribute this to the difficulty of LLMs on the named entity recognition task as reported by other researchers~\cite{Yu2023OpenCO, Ojo2023HowGA}. In comparison to the best-finetuned model, there is a large drop in performance of $-53.2$. This shows that having training data is still relevant for this task even in the LLM era, especially for low-resource languages. 

\paragraph{Large gap in the performance of closed and open models} For intent detection, we find that all open models achieved below 50\% on the relatively easy task of intent detection. The poor performance may be attributed to either a lack of exposure to many African languages or the large label space (i.e. 40) for the classification task. The closed models result are better, with \gpto and \gemini achieving more than $+20$ points than the best open model, Llama 3.3 70B. However, if we compare the results in the English language, open models such as Gemma 2 27B and Llama 3.3 70B are competitive with closed models. This shows that open models are more biased toward high-resource languages. This implies that there is a continuous need to keep improving the capabilities of models for low-resource languages. 

\paragraph{Intent detection performance varies by languages} The performance of some African languages is often higher than others, this is probably connected to the amount of monolingual data available on the web. For example, Swahili (\texttt{swa}) with over 1 billion monolingual data~\citep{kudugunta2023madlad} has $80.6$ accuracy point that is comparable performance to English performance ($81.1$) with Llama 3.3 70B, while other languages have much lower performance. Similarly, \gpto has more than $70$ accuracy points for Amharic, Hausa, Igbo, Shona, Swahili, Xhosa, \yoruba, and Zulu. These languages also have larger monolingual data on the web than the ones with lower than $70\%$ accuracy.



\begin{figure*}
    \centering
    \includegraphics[width=1\linewidth]{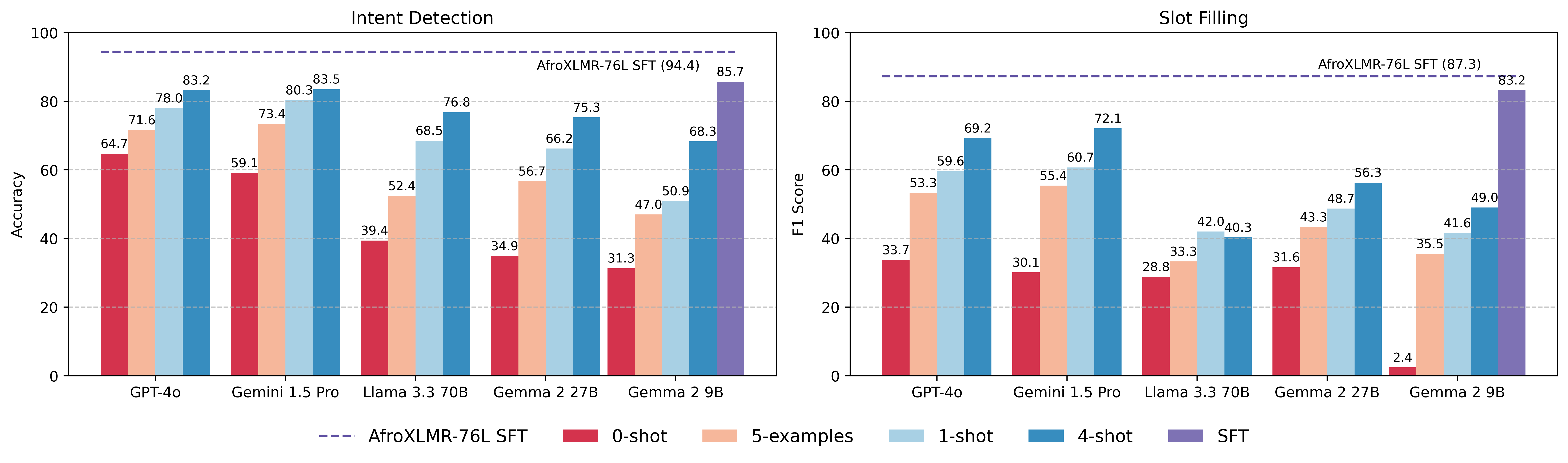}
    \vspace{-8mm}
    \caption{Performance of cross-lingual transfer across different shot settings and supervised fine-tuning (SFT) on the merged 17 languages \dataset{} dataset.}
    \label{fig:shot-sft-comparison}
\end{figure*}

\subsection{Few-shot performance}
\autoref{fig:shot-sft-comparison} shows the result of the various few-shot setups: 5-examples, 1-shot (40 examples, one from each intent type), and 4-shots (160 examples). Our result shows a big boost in performance with only 5-examples, especially for the \textbf{slot-filling task} and some LLMs: \gpto and \gemini improved the most by more than $+19$ points. Similarly, Gemma 2 9B improved from $2.4$ to $33.5$ matching the performance of Llama 3.3 70B (with 5-examples). Additional samples from 1-shot and 4-shot consistently improved performance for all models except Llama 3.3 70B. Similarly, for \textbf{intent detection}, there is consistent improvement in performance with more examples used for few-shot evaluations. We find Gemini 1.5 Pro, Gemma 2 9B and Gemma 2 27B to benefit the most from 5-examples, with an accuracy boost of $+14.3$, $+15.7$, and $+21.8$ respectively. Interestingly,  while the zero-shot performance of Gemini 1.5 Pro is worse than \gpto, the few-shot performance exceeds that of \gpto with $+1.8$ and $+2.3$ improvement in 5-examples and 1-shot.  Our result shows the effectiveness of LLMs in adapting quickly to a new task in low-resource settings. We provide the results of individual languages in Appendix \ref{app:llm-few-shot-results}.

\paragraph{Would Few-shot performance match Supervised fine-tuning (SFT)?} While all LLMs improve performance with more shots, there is still a large gap with SFT. We performed SFT on Gemma 2 9B with all training samples and prompt templates, we found a large performance gap of $+17.4$ and $+34.2$ for intent detection and slot-filling respectively if we compare SFT (52k samples) to 4-shots (160 examples). However, for closed models, the gap of SFT on Gemma 2 9B to \gemini and \gpto is much smaller, especially for intent detection. In general, few-shots of LLMs are still worse than SFT but are very crucial and effective when training data are scarce. 

\begin{figure}
    \centering
    \includegraphics[width=\linewidth]{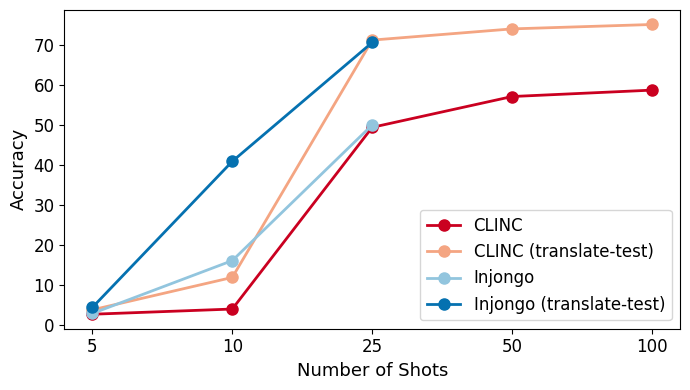}
    \vspace{-7mm}
    \caption{Cross-lingual transfer results from \textsc{CLINC} and \textsc{Injongo} English data}
    \label{fig:cross-lingual-transfer}
    \vspace{-1em}
\end{figure}

\subsection{Cross-lingual Transfer results}
\autoref{fig:cross-lingual-transfer} shows our final experiments that compare cross-lingual transfer results from two English datasets: \textsc{Clinc} (Western-centric) and \dataset{} (African-centric) on the intent detection task. At 5-shots, in both i\textit{n-language} and \textit{translate-test} settings, the accuracy of all settings is quite similar, however as we increase the number of instances to 10-shots (400 examples), we find that the \dataset{} in-language performance was better than the \textsc{Clinc} ($16.1$ vs. $4.0$) that is more Western-centric. Similarly, in \textit{translate-test} setting, the gain in performance is much larger ($+29$)  which implies that in a low-resource setting, leveraging a multicultural dataset with the African context is effective. However, with more samples (25-shots and above), there is no significant difference in whether the samples are Western-centric or not, and training data size seems to be more important.

\section{Conclusion}

We present \dataset{}, a new benchmark dataset for evaluating intent detection and slot-filling for 16 African languages. 
\dataset{} represents the first large-scale multicultural dataset focused on African language Conversation AI. Our experiments reveal that while fine-tuned multilingual models such as AfroXLMR-76L achieved strong performance LLMs still struggle with African languages, particularly in slot filling tasks. 
We hope \dataset{} will accelerate the development of more effective and culturally-aware conversational AI systems for African languages.

\section*{Limitations}
The scope of \dataset{} is constrained by its coverage of only 5 domains and 40 intents, missing some other domains such as healthcare and education that are essential for real-world applications. Our language selection, while substantial, still represents only a fraction of Africa's linguistic diversity, particularly lacking representation from other language families such as Nilo-Saharan languages.
The annotation process revealed inherent challenges in entity classification across languages, requiring two rounds of review to achieve consistent quality. 
Although significant for low-resource languages, the dataset size of 3,200 examples per language remains modest compared to high-resource benchmarks, potentially limiting model performance. Additionally, the fixed distribution of examples across intents may not accurately reflect the natural frequency of these interactions in real-world conversations.


\bibliography{custom,peter_custom}

\newpage
\appendix
\newpage
\section{\dataset{} Dataset}
\label{app:dataset}

\subsection{Statistics of Corpus and Slot Entities}
\label{app:stat}
\begin{table}[!hbt]
  \centering
  \resizebox{0.5\textwidth}{!}{%
  \begin{tabular}{cc|ccc|cccccc}
  \toprule[1pt]
    \multicolumn{2}{c|}{\textbf{Language}} & \multicolumn{3}{c|}{\textbf{Corpus Statistics in Token}} & \multicolumn{6}{c}{\textbf{Slot Entities}} \\
    Code & Name & Total & Avg. & Unique & Total & Avg. & Unique & Un. Fleiss'$\kappa$ & Fleiss'$\kappa$ & $\delta$ \\ \midrule
  \lang{amh} & Amharic & 24233 & 7.573 & 5270 & 10748 & 3.36 & 33 & 0.836 & 0.933 & +0.096 \\
  \lang{ewe} & Ewe & 33210 & 10.378 & 4422 & 11563 & 3.61 & 34 & 0.854 & 1.000 & +0.146 \\
  \lang{hau} & Hausa & 32330 & 10.103 & 1896 & 11792 & 3.69 & 33 & 0.863 & 0.996 & +0.133 \\
  \lang{ibo} & Igbo & 35036 & 10.928 & 3860 & 12639 & 3.94 & 33 & 0.798 & 0.973 & +0.175 \\
  \lang{kin} & Kinyarwanda & 30216 & 9.443 & 6112 & 10753 & 3.36 & 34 & 0.712 & 0.959 & +0.247 \\
  \lang{lin} & Lingala & 29571 & 9.241 & 2672 & 11400 & 3.56 & 33 & 0.798 & 0.990 & +0.192 \\
  \lang{lug} & Luganda & 33368 & 10.418 & 6589 & 12262 & 3.83 & 33 & 0.864 & 0.990 & +0.126 \\
  \lang{orm} & Oromo & 29429 & 9.197 & 5706 & 12570 & 3.93 & 33 & 0.844 & 0.992 & +0.148 \\
  \lang{sna} & Shona & 32901 & 10.282 & 8206 & 15779 & 4.93 & 33 & 0.934 & 0.976 & +0.042 \\
  \lang{sot} & Sotho & 29515 & 9.223 & 3323 & 6699 & 2.09 & 34 & 0.670 & 0.997 & +0.327 \\
  \lang{swa} & Swahili & 38822 & 12.132 & 4603 & 14750 & 4.61 & 34 & 0.864 & 0.985 & +0.121 \\
  \lang{twi} & Twi & 44303 & 13.845 & 4775 & 14881 & 4.65 & 34 & 0.913 & 0.986 & +0.074 \\
  \lang{wol} & Wolof & 37120 & 11.600 & 3460 & 11265 & 3.52 & 33 & 0.726 & 0.941 & +0.215 \\
  \lang{xho} & Xhosa & 26118 & 8.162 & 5086 & 12673 & 3.96 & 33 & 0.804 & 0.936 & +0.132 \\
  \lang{yor} & Yoruba & 43319 & 13.537 & 3103 & 13886 & 4.34 & 34 & 0.847 & 0.988 & +0.141 \\
  \lang{zul} & Zulu & 26496 & 8.285 & 7742 & 12330 & 3.86 & 34 & 0.618 & 0.912 & +0.294 \\
  \lang{eng} & English & 20266 & 10.861 & 3097 & -- & -- & -- & -- & -- & -- \\ \bottomrule[1pt]
  \end{tabular}
  }
  \caption{Statistics of the \dataset{} dataset across 17 languages, including corpus statistics (token counts and distributions) and slot entity analysis (entity counts, averages, and inter-annotator agreement measures) with \textbf{unmerged} slot types.}
  \label{tab:dataset-stats}
\end{table}

\subsection{Categories of Intent Detection}
\label{sec:intent-label}
The following are the intent labels used in the \dataset{} dataset. These are a total of 40 categories across 5 domains (Banking, Kitchen and Dining, Travel, Utility, and Home).

\begin{table}[h!]
\centering
\resizebox{\columnwidth}{!}{%
\begin{tabular}{c|l}
\toprule
\textbf{Domain} & \textbf{Intent} \\ \midrule
 & freeze\_account, pin\_change, pay\_bill, interest\_rate, \\ Banking &  min\_payment, bill\_balance, balance, spending\_history, \\ &  transactions, transfer \\ \hline 
Kitchen & food\_last, confirm\_reservation, ingredients\_list, cook\_time, \\ 
and & restaurant\_reviews, meal\_suggestion, restaurant\_suggestion, \\ 
Dining & restaurant\_reservation, cancel\_reservation, recipe \\ \hline 
Home & play\_music, calendar\_update, update\_playlist, \\ &  shopping\_list\_update \\ \hline 
 & plug\_type, travel\_notification, translate, international\_visa, \\ Travel &  exchange\_rate, travel\_suggestion, book\_flight, book\_hotel, \\ & car\_rental \\ \hline 
Utility & weather, alarm, share\_location, make\_call, time, text \\ \bottomrule 
\end{tabular}
}
\caption{Grouped intents categories by five domains.}
\label{tab:categorized-intents}
\end{table}

\subsection{Categories of Slot Filling}
\label{sec:slot-label}
\autoref{tab:slot-types} shows the original slot types and their final status after merging similar or low-frequency types during preprocessing. The ``Original Slot Type''are used during the dataset annotation phase, which contained 34 slot types.
After merging similar or low-frequency types during data preprocessing in Section \ref{sec:label-merge}, it was reduced to 23 distinct slot types as shown in the ``Final Merged Type'' column.

\begin{table}[h!]
\centering
\resizebox{0.9\columnwidth}{!}{%
\begin{tabular}{l|l|l}
\toprule
\textbf{Original Slot Type} & \textbf{Status} & \textbf{Final Merged Type} \\ \midrule
account type & kept & account type \\
artist name & kept & artist name \\
bank name & kept & bank name \\
bill type & kept & bill type \\
calendar event & kept & calendar event \\
country & kept & country \\
currency & kept & currency \\
date & kept & date \\
hotel name & kept & hotel name \\
language name & kept & language name \\
meal period & kept & meal period \\
money & kept & money \\
music genre & kept & music genre \\
number & kept & number \\
payment company & kept & payment company \\
personal name & kept & personal name \\
place name & kept & place name \\
restaurant name & kept & restaurant name \\
shopping item & kept & shopping item \\
song name & kept & song name \\
time & kept & time \\ \midrule
airline & deleted & -- \\
airport name & deleted & -- \\
car rental company & deleted & -- \\
car type & deleted & -- \\
continent & deleted & -- \\
nationality & deleted & -- \\
plug type & deleted & -- \\
supermarket name & deleted & -- \\
timezone & deleted & -- \\ \midrule
city name & merged & \multirow{2}{*}{city or province} \\
state or province & merged & \\
dish name & merged & \multirow{2}{*}{dish or food} \\
food item & merged & \\ \bottomrule
\end{tabular}
}
\caption{Original and final slot types in the \dataset{} dataset. ``kept'' indicates the slot type was retained, while ``deleted'' indicates the slot type was removed. ``merged'' indicates the slot type was combined with another similar type.}
\label{tab:slot-types}
\end{table}

\section{Experiments Setup}
\label{sec:exp-more}

\subsection{Training Configuration}
To ensure equitable comparison across architectures, we implement a standardized training protocol. All SLMs are finetuned using the AdamW optimizer in 20 epochs with a learning rate schedule incorporating 10\% linear warmup steps followed by linear decay. Early stopping (patience=5) is adopted, and the dev set performance is monitored. Learning rates are carefully calibrated for each architecture type as detailed in \autoref{tab:lr-choice}. Our empirical investigations demonstrate that slot filling tasks consistently require higher learning rates compared to intent detection tasks specifically, encoder-only models utilize $1\times10^{-5}$/$3\times10^{-5}$ for intent detection/slot filling respectively, while encoder-decoder architectures necessitate elevated rates of $5\times10^{-5}$/$1\times10^{-4}$.

Given the computational constraints of finetuning LLMs, Fully Supervised Fine-Tuning (FSFT) is exclusively performed on the Gemma 2 9B model with 5 epochs. Based on established SFT practices and task-specific requirements, we use learning rates of $2\times10^{-5}$ and $2.5\times10^{-5}$ for intent detection and slot filling respectively. Training data is constructed from the combined train splits of \dataset{} dataset across all 17 languages, with prompts randomly sampled from a pool of 5 predefined templates.

All experiments are conducted using full precision (FP32) on NVIDIA H100/L40S GPUs with a consistent batch size of 32, achieved through gradient accumulation when necessary. 

\subsection{Learning Rate Choice}
\label{sec:lr-impact}
Before the final model training, we conducted a comprehensive analysis of learning rate variations to understand their effect on model performance across Intent Detection and Slot Filling tasks. This investigation helped determine optimal learning rates for different model architectures. \autoref{tab:lr-optimize} presents detailed results, extending the findings from \autoref{tab:lr-choice}.

\begin{table}[h]
\centering
\resizebox{0.5\textwidth}{!}{%
\begin{tabular}{l|rcc}
\toprule
Task & \multicolumn{1}{l}{Encoder Only} & \multicolumn{1}{l}{Encoder-Decoder} & \multicolumn{1}{l}{NLLB LLM2Vec} \\ \midrule
Intent Detection & $1\times10^{-5}$ & $5\times10^{-5}$ & $1\times10^{-4}$ \\
Slot Filling & $3\times10^{-5}$ & $1\times10^{-4}$ & $3\times10^{-4}$ \\
\bottomrule
\end{tabular}%
}
\caption{Selected learning rates for different architectures of both tasks of intent detection and slot filling.}
\label{tab:lr-choice}
\end{table}
\vspace{-1em}

\subsection{Language Coverage of Baselines}
\label{app:language}
The table below briefly introduces the baseline models along with the languages they were trained on in the \dataset{} dataset.

\begin{table}[h]
\centering
\resizebox{0.49\textwidth}{!}{%
\begin{tabular}{l|l}
\toprule
\textbf{Model} & \textbf{Languages} \\ \midrule
AfroXLMR Large (550M) & amh, hau, ibo, kin, orm, sna, sot, swa, xho, yor, zul \\ \midrule
AfroXLMR Large 76L (550M) & amh, ewe, hau, ibo, kin, lin, lug, orm, sna, sot, swa, twi, wol, xho, yor, zul \\ \midrule
XLM-RoBERTa Large (550M) & amh, orm, swa, xho \\ \midrule
AfriBERTa V2 Large (187M) & amh, hau, ibo, sna, sot, swa, xho, yor \\ \midrule
AfriTeVa V2 Large (1.2B) & amh, hau, ibo, sna, sot, swa, xho, yor \\ \midrule
mT5-Large (1.2B) & amh, hau, ibo, sot, swa, yor, zul \\ \bottomrule
\end{tabular}
}
\caption{Baseline models with their corresponding language coverage in \dataset{}.}
\label{tab:model_languages}
\end{table}



\begin{table*}[h]
\centering
\resizebox{\textwidth}{!}{%
\begin{tabular}{c|cl|cccccccc}
\toprule[1pt]
\multirow{2}{*}{\textbf{Task}} & \multirow{2}{*}{\begin{tabular}[c]{@{}c@{}}\textbf{Model}\\\textbf{Type}\end{tabular}} & \multirow{2}{*}{\textbf{Model}} & \multicolumn{8}{c}{\textbf{Learning Rate}} \\
  &  &  & $1\times10^{-5}$ & $2\times10^{-5}$ &$3\times10^{-5}$& $5\times10^{-5}$& $1\times10^{-4}$& $2\times10^{-4}$ & $3\times10^{-4}$ & $5\times10^{-4}$ \\ \midrule
\multirow{10}{*}{\begin{tabular}[c]{@{}c@{}}\textsc{Intent}\\\textsc{Detection}\end{tabular}} & \multirow{5}{*}{Encoder} & AfriBERTa V2 Large & 97.50 & 98.13 & 98.13 & 98.13 & 97.81 & 97.81 & 95.00 & 2.50 \\
  &  & AfroXLMR-large & 98.13 & 98.13 & 98.75 & 2.50 & 2.50 & 2.50 & 2.50 & 2.50 \\
  &  & AfroXLMR-large 76L & 98.75 & 98.75 & 99.06 & 98.75 & 2.50 & 2.50 & 2.50 & 2.50 \\
  &  & XLM-RoBERTa Large & 98.75 & 2.50 & 13.44 & 6.25 & 2.50 & 2.50 & 4.06 & 2.50 \\
  &  & \textit{Average} & \textbf{98.28} & 74.38 & 77.34 & 51.41 & 26.33 & 26.33 & 26.02 & 2.50 \\ \cline{2-11} 
  & \multirow{3}{*}{\begin{tabular}[c]{@{}c@{}}Encoder-\\Decoder\end{tabular}} & AfriTeVa V2 Large & 0.00 & 0.00 & 96.88 & 97.81 & 97.19 & 97.81 & 96.56 & 97.50 \\
  &  & mT5-Large & 0.00 & 95.31 & 95.94 & 97.50 & 97.50 & 97.50 & 98.13 & 97.50 \\
  &  & \textit{Average} & 0.00 & 47.66 & 96.41 & \textbf{97.66} & 97.34 & \textbf{97.66} & 97.34 & 97.50 \\ \cline{2-11} 
  & \multirow{2}{*}{Other} & NLLB LLM2Vec & 97.19 & 97.50 & 96.88 & 95.94 & 98.44 & 97.81 & 98.44 & 97.19 \\
  &  & \textit{Average} & 97.19 & 97.50 & 96.88 & 95.94 & \textbf{98.44} & 97.81 & \textbf{98.44} & 97.19 \\ \midrule
\multirow{10}{*}{{\begin{tabular}[c]{@{}c@{}}\textsc{Slot}\\\textsc{Filling}\end{tabular}}} & \multirow{5}{*}{Encoder} & AfriBERTa V2 Large & 86.12 & 89.70 & 90.21 & 90.74 & 91.22 & 90.45 & 88.24 & 0.00 \\
  &  & AfroXLMR-large & 89.95 & 90.13 & 91.04 & 89.87 & 0.00 & 0.00 & 0.00 & 0.00 \\
  &  & AfroXLMR-large 76L & 90.04 & 90.91 & 90.96 & 90.58 & 91.28 & 0.00 & 0.00 & 0.00 \\
  &  & XLM-RoBERTa Large & 88.55 & 89.72 & 91.63 & 89.52 & 88.10 & 0.00 & 0.00 & 0.00 \\
  &  & \textit{Average} & 88.67 & 90.11 & \textbf{90.96} & 90.18 & 67.65 & 22.61 & 22.06 & 0.00 \\ \cline{2-11}  
  & \multirow{3}{*}{\begin{tabular}[c]{@{}c@{}}Encoder-\\Decoder\end{tabular}} & AfriTeVa V2 Large & 39.07 & 83.47 & 83.47 & 89.63 & 90.51 & 81.59 & 88.11 & 88.44 \\
  &  & mT5-Large & 22.31 & 59.61 & 89.16 & 82.71 & 88.54 & 89.67 & 89.16 & 90.40 \\
  &  & \textit{Average} & 30.69 & 71.54 & 86.32 & 86.17 & \textbf{89.53} & 85.63 & 88.64 & 89.42 \\ \cline{2-11} 
  & \multirow{2}{*}{Other} & NLLB LLM2Vec & 81.13 & 84.84 & 85.33 & 85.69 & 85.57 & 84.44 & 87.02 & 86.12 \\
  &  & \textit{Average} & 81.13 & 84.84 & 85.33 & 85.69 & 85.57 & 84.44 & \textbf{87.02} & 86.12 \\ 
\bottomrule[1pt]
\end{tabular}%
}
\caption{Comparative analysis of model performance across different learning rates for Intent Detection and Slot Filling tasks. Results are shown for various model architectures including Encoder-only, Encoder-Decoder, and other approaches. Bold values indicate the best performance for each model type.}
\label{tab:lr-optimize}
\end{table*}

\subsection{Results of Multi-lingual Training}
\label{app:multi-lingual-results}
\begin{table*}[h]
\centering
\resizebox{\textwidth}{!}{%
\begin{tabular}{c|l|cccccccccccccccccc}
\toprule[1pt]
\multicolumn{1}{l|}{\textbf{Task}} & \textbf{Model} & \lang{amh} & \lang{ewe} & \lang{hau} & \lang{ibo} & \lang{kin} & \lang{lin} & \lang{lug} & \lang{orm} & \lang{sna} & \lang{sot} & \lang{swa} & \lang{twi} & \lang{wol} & \lang{xho} & \lang{yor} & \lang{zul} & \lang{eng} & \textbf{AVG} \\ \midrule
  \multirow{4}{*}{\begin{tabular}[c]{@{}c@{}}\textsc{Intent} \\ \textsc{Detection}\end{tabular}} 
  & AfroXLMR-large 76L& 96.0& 92.6& 99.2& 96.6& 87.7& 95.9& 92.3& 92.9& 96.5& 87.6& 97.8& 94.2& 97.1& 97.3& 97.9& 89.2& 89.0& \textbf{94.4$_{\pm 3.6}$} \\
  & AfroXLMR-large& 96.1& 90.3& 99.3& 96.5& 86.8& 94.2& 91.6& 92.2& 96.0& 87.1& 97.9& 91.6& 96.1& 96.9& 97.4& 88.6& 89.7& 93.7$_{\pm 3.9}$ \\
  & NLLB LLM2Vec& 95.8& 90.2& 98.7& 96.5& 86.2& 95.4& 92.6& 87.9& 96.9& 86.8& 97.3& 93.9& 95.5& 96.9& 97.2& 88.6& 89.1& 93.5$_{\pm 4.1}$ \\
  & AfriTeVa V2 Large& 94.6& 85.8& 99.2& 96.5& 87.3& 93.6& 90.8& 88.6& 95.9& 85.3& 98.0& 89.6& 94.4& 97.3& 97.6& 88.3& 89.7& 92.7$_{\pm 4.6}$ \\ \midrule
  \multirow{4}{*}{{\begin{tabular}[c]{@{}c@{}}\textsc{Slot} \\\textsc{Filling}\end{tabular}}} 
  & AfroXLMR-large 76L& 88.2& 87.0& 96.3& 84.0& 79.3& 90.3& 89.2& 87.2& 86.1& 80.4& 90.5& 90.3& 83.3& 91.8& 90.2& 83.3& 82.4& \textbf{87.3$_{\pm 4.4}$} \\
  & AfroXLMR-large& 87.9& 84.0& 96.4& 83.6& 80.4& 89.5& 88.4& 88.2& 87.0& 82.0& 91.5& 87.7& 81.9& 91.7& 90.4& 84.2& 82.8& 87.2$_{\pm 4.2}$ \\
  & NLLB LLM2Vec& 84.3& 82.0& 94.6& 80.3& 72.3& 86.9& 85.1& 81.9& 82.0& 77.2& 87.3& 85.8& 80.0& 90.4& 87.1& 79.9& 80.8& 83.6$_{\pm 5.2}$ \\
  & AfriTeVa V2 Large& 78.9& 72.6& 92.0& 80.0& 75.7& 85.3& 81.8& 76.0& 79.8& 77.0& 88.2& 81.7& 76.5& 86.7& 86.3& 66.7& 78.9& 80.3$_{\pm 6.3}$ \\ \bottomrule[1pt]
\end{tabular}%
}
\caption{Multilingual Training: 4 model performance on Intent Detection and Slot Filling tasks across languages.}
\label{tab:combined-languages-sft-result}
\end{table*}

We selected the 4 top-performing models from the in-language training phase and evaluated them on the \dataset{} test set, comparing the performance of the models when trained on individual languages and when trained on the combined dataset. The results are shown in \autoref{tab:combined-languages-sft-result}.


\subsection{Results of Cross-lingual Transfer}

This section provides additional commentary on \autoref{tab:afroxlm76l_intent_detection} which reports the cross-lingual transfer performance of AfroXLMR-76L on the Intent Detection task under different shot conditions. The table compares two datasets (CLINC and Injongo) in both their original in-language and translate-test settings. For each dataset, results are presented at multiple shot levels (e.g., 5, 10, 25, 50, and 100 shots), with the average performance and corresponding standard deviation indicated. Notably, the results illustrate how performance progressively improves as the number of shots increases, and how the transfer capability is affected by the linguistic diversity of the datasets.

\begin{table*}[htbp]
\centering
\resizebox{\textwidth}{!}{%
\begin{tabular}{c*{17}{c}}
\toprule
\textbf{\# of Shots} & \lang{amh} & \lang{ewe} & \lang{hau} & \lang{ibo} & \lang{kin} & \lang{lin} & \lang{lug} & \lang{orm} & \lang{sna} & \lang{sot} & \lang{swa} & \lang{twi} & \lang{wol} & \lang{xho} & \lang{yor} & \lang{zul} & \textbf{AVG} \\
\midrule
\multicolumn{18}{l}{\textbf{CLINC Dataset}} \\
5 & 3.1 & 2.7 & 2.5 & 2.6 & 2.5 & 3.2 & 2.6 & 2.4 & 2.9 & 2.9 & 3.6 & 1.9 & 2.8 & 2.3 & 2.7 & 3.0 & 2.7$_{\pm 0.5}$ \\
10 & 6.7 & 2.8 & 5.5 & 4.1 & 3.5 & 5.1 & 3.2 & 3.2 & 3.1 & 4.2 & 7.0 & 2.7 & 3.1 & 2.6 & 3.4 & 3.6 & 4.0$_{\pm1.0}$ \\
25 & 80.3 & 24.4 & 69.6 & 51.9 & 49.9 & 57.2 & 37.6 & 29.8 & 53.0 & 42.9 & 78.0 & 38.4 & 26.4 & 59.6 & 35.6 & 55.1 & 49.4$_{\pm9.4}$ \\
50 & 83.8 & 36.1 & 78.4 & 61.9 & 55.3 & 63.4 & 45.7 & 39.4 & 59.8 & 50.3 & 82.6 & 47.1 & 34.9 & 65.3 & 48.7 & 60.0 & 57.1$_{\pm8.4}$ \\
100 & 84.7 & 37.8 & 80.9 & 62.6 & 55.7 & 65.2 & 47.2 & 39.6 & 63.2 & 50.9 & 85.2 & 48.8 & 36.0 & 66.7 & 52.5 & 62.1 & 58.7$_{\pm8.5}$ \\
\midrule
\multicolumn{18}{l}{\textbf{Injongo Dataset}} \\
5 & 3.6 & 2.6 & 3.3 & 2.6 & 3.1 & 3.0 & 2.8 & 2.3 & 3.2 & 3.3 & 3.9 & 2.2 & 2.8 & 2.8 & 2.5 & 2.8 & 2.9$_{\pm0.5}$ \\
10 & 29.7 & 7.3 & 27.0 & 15.3 & 13.9 & 19.4 & 9.3 & 6.3 & 13.4 & 16.2 & 36.7 & 9.4 & 7.1 & 16.5 & 10.0 & 20.3 & 16.1$_{\pm5.1}$ \\
25 & 76.1 & 24.2 & 70.2 & 53.9 & 50.1 & 55.2 & 41.2 & 28.4 & 53.9 & 45.1 & 78.5 & 41.1 & 27.8 & 59.7 & 38.8 & 55.1 & 50.0$_{\pm8.9}$ \\
\midrule
\multicolumn{18}{l}{\textbf{CLINC Dataset (translate  test)}} \\
5 & 4.2 & 3.6 & 3.5 & 4.0 & 3.9 & 4.0 & 3.6 & 3.5 & 3.7 & 3.8 & 4.3 & 3.2 & 3.6 & 3.9 & 3.6 & 4.0 & 3.8$_{\pm0.5}$ \\
10 & 14.9 & 11.2 & 15.9 & 13.7 & 11.9 & 15.3 & 11.0 & 4.8 & 10.2 & 12.4 & 13.8 & 9.4 & 10.2 & 12.4 & 10.3 & 12.5 & 11.9$_{\pm1.9}$ \\
25 & 83.2 & 58.8 & 85.7 & 78.3 & 67.6 & 76.7 & 71.4 & 32.1 & 80.2 & 67.4 & 84.6 & 67.8 & 56.8 & 82.8 & 75.3 & 71.0 & 71.2$_{\pm7.4}$ \\
50 & 86.2 & 61.7 & 89.9 & 81.9 & 69.1 & 78.8 & 74.2 & 33.2 & 82.7 & 71.7 & 85.8 & 70.1 & 58.1 & \textbf{86.7} & 79.2 & 74.4 & 74.0$_{\pm7.7}$ \\
100 & \textbf{86.7} & \textbf{62.5} & \textbf{91.0} & \textbf{83.3} & \textbf{69.5} & \textbf{80.5} & \textbf{75.4} & \textbf{34.4} & \textbf{84.9} & \textbf{72.2} & \textbf{87.7} & \textbf{71.1 }& \textbf{59.1} & 86.2 & \textbf{81.2} & \textbf{76.6} & \textbf{75.1}$_{\pm7.7}$ \\
\midrule
\multicolumn{18}{l}{\textbf{Injongo Dataset (translate test)}} \\
5 & 4.4 & 4.7 & 5.3 & 4.4 & 3.7 & 5.0 & 4.4 & 2.7 & 4.5 & 4.5 & 4.7 & 3.8 & 4.2 & 5.1 & 5.3 & 3.6 & 4.4$\pm$0.6 \\
10 & 45.4 & 32.2 & 52.0 & 44.7 & 39.7 & 43.9 & 42.8 & 16.3 & 44.2 & 38.6 & 50.6 & 37.1 & 32.7 & 48.0 & 45.3 & 40.8 & 40.9$\pm$4.9 \\
25 & 80.5 & 59.0 & 86.0 & 77.5 & 66.4 & 74.2 & 72.7 & 31.7 & 77.9 & 68.1 & 84.1 & 66.6 & 55.9 & 82.4 & 76.8 & 69.7 & 70.6$\pm$7.3 \\
\bottomrule
\end{tabular}%
}
\caption{AfroXLMR-76L Intent Detection performance under varying shot conditions.}
\label{tab:afroxlm76l_intent_detection}
\end{table*}

\subsection{Inference Setup of LLMs}

For closed‐source models (GPT-4o and Gemini 1.5 Pro), we utilize the API provided by the respective vendor for inference. For open‐source models, inference is performed using vLLM \cite{vllm}, except for Aya-101, where Text Generation Inference (TGI)\footnote{\href{https://huggingface.co/docs/text-generation-inference/index}{Text Generation Inference}} is employed.


\subsection{Results of LLMs prompting}
\label{app:llm-few-shot-results}
Across 5 LLMs, we evaluated the performance of zero-shot and few-shot learning on the Intent Detection and Slot Filling tasks. The complete results are presented in \autoref{tab:few-shot-results}. We only evaluate the performance of the models on the best prompt for each task. The 2nd prompt for Intent Detection and the 3rd prompt for Slot Filling are used for evaluation. 

\begin{table*}[htbp!]
\centering
\resizebox{\textwidth}{!}{%
\begin{tabular}{c|l|l|cccccccccccccccccc}
\toprule
\textbf{Task} & \textbf{Model} & \textbf{Setup} & \lang{eng} & \lang{amh} & \lang{ewe} & \lang{hau} & \lang{ibo} & \lang{kin} & \lang{lin} & \lang{lug} & \lang{orm} & \lang{sna} & \lang{sot} & \lang{swa} & \lang{twi} & \lang{wol} & \lang{xho} & \lang{yor} & \lang{zul} & \textbf{Avg} \\ \midrule
\multirow{20}{*}{\begin{tabular}[c]{@{}c@{}} \textsc{Intent} \\ \textsc{Detection}\end{tabular}} & \multirow{4}{*}{GPT-4o} & 0 shot & 81.2 & 76.2 & 14.5 & 80.8 & 71.6 & 64.4 & 55.9 & 68.1 & 58.6 & 75.6 & 58.6 & 85.2 & 58.3 & 43.1 & 78.6 & 76.1 & 70.3 & 64.7$_{\pm 16.8}$ \\
& & 5 examples & 81.8 & 75.9 & 21.2 & 85.3 & 76.6 & 69.8 & 74.8 & 74.7 & 69.1 & 80.8 & 69.2 & 82.2 & 68.9 & 63.3 & 82.0 & 78.4 & 72.7 & 71.6$_{\pm 14.2}$ \\
& & 1 shot & 82.6 & 83.0 & 37.8 & 88.8 & 82.0 & 76.2 & 82.2 & 83.3 & 78.6 & 85.0 & 71.2 & 85.8 & 76.7 & 72.6 & 85.2 & 82.0 & 77.3 & 78.0$_{\pm 11.4}$ \\
& & 4 shots & 83.3 & 85.2 & 64.5 & \textbf{91.7} & \textbf{86.7} & \textbf{79.4} & 85.0 & \textbf{85.3} & \textbf{83.3} & \textbf{89.7} & \textbf{75.0} & 87.8 & 82.2 & 81.1 & \textbf{87.2} & \textbf{87.7} & 79.2 & 83.2$_{\pm 6.4}$ \\ \cmidrule{2-21}
& \multirow{4}{*}{Gemini 1.5 Pro} & 0 shot & 82.3 & 80.2 & 26.9 & 78.8 & 69.5 & 66.2 & 58.1 & 64.4 & 42.8 & 71.4 & 55.5 & 85.0 & 50.5 & 27.9 & 79.4 & 71.6 & 71.9 & 62.5$_{\pm 17.2}$ \\
& & 5 examples & 81.8 & 81.1 & 52.3 & 86.4 & 77.3 & 71.4 & 76.4 & 76.1 & 67.0 & 80.2 & 69.2 & 85.5 & 71.2 & 49.1 & 81.1 & 77.3 & 72.8 & 73.4$_{\pm 10.1}$ \\
& & 1 shot & 81.2 & 85.8 & 69.2 & 90.2 & 80.3 & 75.6 & 82.5 & 82.8 & 74.7 & 85.2 & 73.3 & 87.3 & 80.8 & 71.2 & 84.7 & 84.5 & 77.5 & 80.3$_{\pm 5.9}$ \\
& & 4 shots & \textbf{83.8} & \textbf{85.9} & \textbf{78.3} & 90.9 & 86.2 & 79.1 & \textbf{85.6} & 83.6 & 78.0 & 87.7 & 73.6 & \textbf{88.9} & \textbf{84.2} & \textbf{81.6} & 86.9 & 85.6 & \textbf{80.3} & \textbf{83.5$_{\pm 4.5}$} \\ \cmidrule{2-21}
& \multirow{4}{*}{Gemma 2 IT 9B} & 0 shot & 78.9 & 51.4 & 7.0 & 43.1 & 33.4 & 26.1 & 24.5 & 25.6 & 8.6 & 30.5 & 21.2 & 73.1 & 23.1 & 14.9 & 44.4 & 33.8 & 40.5 & 31.3$_{\pm 16.2}$ \\
& & 5 examples & 79.1 & 58.3 & 13.4 & 71.2 & 58.9 & 44.1 & 41.6 & 40.0 & 18.3 & 54.1 & 41.4 & 79.1 & 39.2 & 29.8 & 61.1 & 48.4 & 53.3 & 47.0$_{\pm 17.0}$ \\
& & 1 shot & 78.9 & 54.7 & 15.5 & 76.7 & 58.8 & 43.3 & 50.3 & 42.2 & 28.4 & 54.7 & 43.1 & 82.0 & 46.7 & 30.3 & 68.3 & 60.5 & 58.8 & 50.9$_{\pm 16.9}$ \\
& & 4 shots & 78.5 & 68.6 & 44.8 & 84.7 & 73.4 & 60.9 & 72.0 & 70.0 & 55.0 & 75.2 & 60.3 & 82.8 & 66.4 & 66.9 & 77.0 & 67.7 & 67.7 & 68.3$_{\pm 9.7}$ \\ \cmidrule{2-21}
& \multirow{4}{*}{Gemma 2 IT 27B} & 0 shot & 80.2 & 48.4 & 6.6 & 49.8 & 40.2 & 27.8 & 31.6 & 28.6 & 6.4 & 38.0 & 27.3 & 77.5 & 23.0 & 18.7 & 51.7 & 35.5 & 47.3 & 34.9$_{\pm 17.5}$ \\
& & 5 examples & 78.3 & 59.5 & 13.6 & 80.0 & 70.0 & 52.2 & 55.5 & 52.5 & 22.5 & 68.3 & 55.3 & 84.7 & 54.5 & 37.0 & 73.9 & 63.3 & 64.2 & 56.7$_{\pm 18.5}$ \\
& & 1 shot & 80.5 & 61.6 & 26.2 & 85.2 & 74.2 & 59.8 & 68.3 & 65.9 & 49.2 & 77.8 & 59.5 & 86.7 & 63.7 & 58.4 & 77.0 & 75.9 & 68.8 & 66.2$_{\pm 14.3}$ \\
& & 4 shots & 81.5 & 76.4 & 57.7 & 87.7 & 80.6 & 65.9 & 78.1 & 74.1 & 65.3 & 83.4 & 68.1 & 85.3 & 74.7 & 70.9 & 81.2 & 80.2 & 75.0 & 75.3$_{\pm 7.9}$ \\ \cmidrule{2-21}
& \multirow{4}{*}{Llama 3.3 70B} & 0 shot & 80.9 & 57.3 & 10.5 & 53.3 & 53.0 & 35.5 & 38.0 & 39.7 & 13.8 & 32.8 & 31.7 & 81.4 & 31.7 & 21.0 & 44.7 & 41.6 & 44.8 & 39.4$_{\pm 16.8}$ \\
& & 5 examples & 82.3 & 56.6 & 12.0 & 79.2 & 69.1 & 51.1 & 45.6 & 48.3 & 28.9 & 51.1 & 47.7 & 84.2 & 50.2 & 31.0 & 62.2 & 63.7 & 58.0 & 52.4$_{\pm 17.7}$ \\
& & 1 shot & 82.2 & 75.3 & 37.0 & 84.7 & 77.8 & 59.2 & 59.7 & 72.8 & 54.1 & 72.3 & 61.3 & 86.7 & 69.8 & 60.3 & 76.4 & 77.8 & 70.3 & 68.5$_{\pm 12.2}$ \\
& & 4 shots & 83.3 & 81.4 & 57.0 & 88.1 & 83.6 & 69.7 & 75.8 & 77.7 & 65.8 & 81.4 & 68.6 & 88.0 & 77.0 & 74.5 & 80.3 & 84.8 & 74.8 & 76.8$_{\pm 8.1}$ \\ \midrule \midrule
\multirow{20}{*}{{\begin{tabular}[c]{@{}c@{}}\textsc{Slot} \\\textsc{Filling}\end{tabular}}} & \multirow{4}{*}{GPT-4o} & 0 shot & 55.1 & 23.8 & 20.3 & 38.8 & 38.9 & 37.3 & 33.6 & 37.9 & 12.7 & 41.4 & 42.6 & 44.5 & 39.0 & 41.3 & 9.1 & 41.7 & 36.9 & 33.7$_{\pm 10.7}$ \\
& & 5 examples & 63.9 & 39.3 & 43.5 & 60.8 & 59.8 & 46.8 & 61.0 & 51.0 & 36.6 & 60.6 & 58.8 & 62.4 & 61.8 & 58.5 & 50.7 & 59.4 & 40.8 & 53.3$_{\pm 8.8}$ \\
& & 1 shot & 71.3 & 53.3 & 50.0 & 66.2 & 59.9 & 54.3 & 63.3 & 60.3 & 54.9 & 64.7 & 56.7 & 67.4 & 61.5 & 56.3 & 67.3 & 67.8 & 50.4 & 59.6$_{\pm 5.9}$ \\
& & 4 shot & \textbf{75.4} & 64.2 & 57.2 & 71.1 & 70.6 & \textbf{62.8} & 74.0 & 74.1 & \textbf{66.8} & 71.3 & 63.5 & 77.1 & 74.4 & 68.4 & 75.8 & 77.8 & \textbf{58.6} & 69.2$_{\pm 6.3}$ \\ \cmidrule{2-21}
& \multirow{4}{*}{Gemini 1.5 Pro} & 0 shot & 48.4 & 20.3 & 18.0 & 30.2 & 34.5 & 33.3 & 33.2 & 34.7 & 14.4 & 33.8 & 40.4 & 33.7 & 34.3 & 33.1 & 7.9 & 35.9 & 36.5 & 29.6$_{\pm 8.9}$ \\
& & 5 examples & 64.7 & 52.6 & 42.3 & 61.3 & 59.2 & 47.7 & 56.8 & 63.1 & 36.5 & 65.6 & 62.4 & 66.1 & 61.8 & 55.4 & 46.1 & 59.6 & 49.7 & 55.4$_{\pm 8.5}$ \\
& & 1 shot & 64.8 & 62.1 & 51.0 & 67.3 & 61.5 & 52.1 & 61.5 & 66.2 & 47.2 & 66.0 & 57.2 & 70.4 & 68.4 & 56.0 & 64.8 & 67.3 & 52.4 & 60.7$_{\pm 7.0}$ \\
& & 4 shots & 75.2 & \textbf{69.0} & \textbf{67.6} & \textbf{72.2} & \textbf{73.4} & 62.5 & \textbf{77.4} & \textbf{77.4} & 66.6 & \textbf{77.0} & \textbf{65.9} & \textbf{79.9} & \textbf{77.2} & \textbf{69.8} & \textbf{80.0} & \textbf{81.0} & 57.4 & \textbf{72.1$_{\pm 6.7}$} \\ \cmidrule{2-21}
& \multirow{4}{*}{Gemma 2 IT 9B} & 0 shot & 27.0 & 0.3 & 0.0 & 3.2 & 5.6 & 1.4 & 3.9 & 2.1 & 0.0 & 2.4 & 2.6 & 13.7 & 0.2 & 0.5 & 0.0 & 0.2 & 3.0 & 2.4$_{\pm 3.3}$ \\
& & 5 examples & 55.0 & 26.4 & 20.5 & 39.8 & 40.3 & 29.7 & 32.0 & 37.7 & 11.9 & 42.7 & 31.5 & 57.1 & 50.0 & 36.6 & 33.8 & 37.5 & 41.0 & 35.5$_{\pm 10.4}$ \\
& & 1 shot & 55.6 & 37.5 & 26.2 & 50.6 & 44.0 & 38.4 & 34.2 & 44.8 & 20.5 & 46.4 & 37.6 & 59.3 & 47.1 & 41.3 & 50.2 & 49.7 & 38.2 & 41.6$_{\pm 9.3}$ \\
& & 4 shots & 59.6 & 45.1 & 38.8 & 61.3 & 51.1 & 41.8 & 47.8 & 51.4 & 32.2 & 56.2 & 37.9 & 65.6 & 51.6 & 49.4 & 58.7 & 54.5 & 41.0 & 49.0$_{\pm 8.9}$ \\ \cmidrule{2-21}
& \multirow{4}{*}{Gemma 2 IT 27B} & 0 shot & 54.0 & 24.1 & 21.9 & 34.8 & 38.3 & 32.4 & 25.6 & 37.6 & 5.2 & 37.8 & 42.1 & 44.7 & 37.8 & 38.2 & 5.9 & 39.4 & 39.4 & 31.6$_{\pm 11.6}$ \\
& & 5 examples & 58.4 & 37.4 & 32.3 & 54.1 & 53.6 & 37.2 & 39.2 & 48.0 & 22.0 & 52.2 & 21.8 & 64.6 & 51.9 & 48.3 & 40.2 & 49.5 & 40.5 & 43.3$_{\pm 11.3}$ \\
& & 1 shot & 41.7 & 21.8 & 37.4 & 60.8 & 58.6 & 45.3 & 45.9 & 54.8 & 27.1 & 55.5 & 46.0 & 67.0 & 56.0 & 49.3 & 48.4 & 58.4 & 47.1 & 48.7$_{\pm 11.6}$ \\
& & 4 shots & 69.8 & 47.2 & 45.0 & 64.4 & 61.5 & 46.8 & 58.3 & 60.8 & 38.8 & 64.5 & 48.6 & 70.8 & 62.2 & 56.2 & 69.3 & 63.3 & 42.7 & 56.3$_{\pm 9.7}$ \\ \cmidrule{2-21}
& \multirow{4}{*}{Llama 3.3 70B} & 0 shot & 55.2 & 28.7 & 24.2 & 28.2 & 35.6 & 30.9 & 22.7 & 32.0 & 11.4 & 29.2 & 31.5 & 43.2 & 34.3 & 37.2 & 7.3 & 33.2 & 30.7 & 28.8$_{\pm 8.7}$ \\
& & 5 examples & 54.0 & 30.7 & 9.5 & 33.2 & 49.6 & 32.1 & 36.7 & 41.5 & 23.5 & 45.2 & 0.0 & 54.4 & 27.6 & 28.5 & 38.7 & 43.3 & 37.7 & 33.3$_{\pm 13.5}$ \\
& & 1 shot & 61.7 & 32.7 & 29.3 & 38.6 & 52.6 & 33.7 & 36.9 & 44.1 & 28.6 & 45.8 & 28.8 & 64.0 & 51.7 & 43.7 & 53.0 & 53.6 & 35.5 & 42.0$_{\pm 10.4}$ \\
& & 4 shots & 62.3 & 35.6 & 20.5 & 28.0 & 32.4 & 36.7 & 53.5 & 38.7 & 34.8 & 40.4 & 22.6 & 63.2 & 53.3 & 46.5 & 45.3 & 59.6 & 32.9 & 40.3$_{\pm 12.1}$ \\
\bottomrule
\end{tabular}
}
\caption{Zero-shot and few-shot performance comparison across languages on and tasks. For Intent Detection, shots refer to examples per domain 5 examples, per (1 shot), and 4 examples per (4 shots). For Slot Filling, shots refer to examples per domain 5 examples, per slot type (1 shot), and 4 examples per slot type (4 shots).}
\label{tab:few-shot-results}
\end{table*}

\newpage
\section{Prompts for Large Language Models}
\label{app:prompts}

We provide the prompts in Jinja format \footnote{\href{https://jinja.palletsprojects.com/en/stable/}{Jinjia: A fast, expressive, extensible templating engine.}} used for the Intent Detection and Slot Filling tasks in the zero-shot and few-shot learning experiments. The prompts are designed to guide the model to perform the specific task on the given input text. The prompts are language-specific and tailored to the task requirements. The prompts 

The variables in the prompts are replaced with the actual input text during the model evaluation. Here is the list of variables used in the prompts:

\begin{itemize}
    \item \texttt{shot\_count}: The number of examples provided to the model, if \texttt{shot\_count} is 0 zero, means zero-shot.
    \item \texttt{examples}: A list of examples provided to the model for few-shot learning.
    \item \texttt{text}: The sentence for which the model needs to predict the intent or slot.
\end{itemize}

\subsection{Intent Detection}
\texttt{\textbf{Prompt I}}
\begin{minted}[breaklines=true, bgcolor=promptColor1]{jinja}
Classify the given sentence by identifying its intent and selecting the most appropriate category from the provided list.

# Steps
1. Analyze the sentence to understand its primary intention or purpose.
2. Compare the identified intention against the possible intent categories.
3. Select the category that best matches the sentence's intent.

# Output Format
- Return the only one matching intent category from the list above. 
- No additional text or punctuation should be included in the output. 
\end{minted}
\texttt{\textbf{Prompt II}}
\begin{minted}[breaklines=true,bgcolor=promptColor2]{jinja}
Identify the intent of the provided text by selecting the most suitable category from the list of available options.

# Steps
1. Analyze the sentence to determine its primary purpose or intention.
2. Match the identified intention with the available intent categories.
3. Choose the category that best aligns with the sentence's intent.

# Output Format
- Return the selected intent category from the list above.
- Do not include any additional text or punctuation in the response.
\end{minted}
\texttt{\textbf{Prompt III}}
\begin{minted}[breaklines=true,bgcolor=promptColor3]{jinja}
Determine the intent of the provided text by selecting the most appropriate category from the given options.

# Steps
1. **Read the Text**: Carefully read the provided text to understand the context and main message.
2. **Identify Key Elements**: Identify the main action, subject, and any relevant details that indicate the overall purpose of the text.
3. **Consider Categories**: Review the list of available categories and consider which category best matches the text's intent.
4. **Reasoning**: Consider why you believe the text fits a certain category by assessing how the identified key elements align with the category's definition.
5. **Selection**: Select the category that most accurately represents the intent of the text.

# Output Format
- Provide the selected category as a plain text response. 
- Don't include any justification.
\end{minted}
\texttt{\textbf{Prompt IV}}
\begin{minted}[breaklines=true,bgcolor=promptColor4]{jinja}
Identify the intent of the provided text by selecting the most suitable category from the list of available options.

# Steps

1. Analyze the text to understand its primary purpose and context.
2. Consider the range of possible intents that the text might express, such as inquiry, statement, request, etc.
3. Match the text with the most appropriate category based on its content and purpose.

# Output Format
Provide the resulting intent category as a short, concise phrase or word that best represents the text's purpose from the available options.

# Notes
- Carefully evaluate any subtleties in the language to determine the intent accurately.
- Consider edge cases where texts might have multiple overlapping intents, and choose the most dominant one.
\end{minted}
\texttt{\textbf{Prompt V}}
\begin{minted}[breaklines=true,bgcolor=promptColor5]{jinja}
Identify the intent of the provided text by selecting the most suitable category from the list of available options.

Consider the subtleties in language and any overlapping intents to determine the most dominant intent category.

# Steps

1. **Analyze the Text**: Thoroughly read and understand the text to grasp its primary purpose and context.
2. **Consider Possible Intents**: Reflect on the range of potential intents the text could express, such as inquiry, statement, or request.
3. **Match with Category**: Align the text with the most appropriate category based on content, language subtleties, and dominant purpose.

# Output Format

Provide the resulting intent category as a short, concise phrase or word.

# Notes

- Pay attention to context and subtleties in the text.
- Evaluate texts with multiple intents, prioritizing the most dominant one.
\end{minted}
\texttt{\textbf{Suffix for Zero-shot and Few-shot}}

\begin{minted}[breaklines=true]{jinja}
# Intent Categories
alarm, balance, bill_balance, book_flight, book_hotel, calendar_update, cancel_reservation, car_rental, confirm_reservation, cook_time, exchange_rate, food_last, freeze_account, ingredients_list, interest_rate, international_visa, make_call, meal_suggestion, min_payment, pay_bill, pin_change, play_music, plug_type, recipe, restaurant_reservation, restaurant_reviews, restaurant_suggestion, share_location, shopping_list_update, spending_history, text, time, timezone, transactions, transfer, translate, travel_notification, travel_suggestion, update_playlist, weather

{% if shot_count == 0 -%}
{# Zero-shot Suffix #}
# Format Example:
Sentence: Can you tell me the weather forecast for today?
Output: weather
{% else %}
{# Few-shot Suffix #}
{% for example in examples -%}
Sentence: {{ example.text }}
Output: {{ example.intent }}

{% endfor %}
Based on the example, consider the following:
{% endif %}
Sentence: {{ text }}
Output: 
\end{minted}


\subsection{Slot Filling}
\texttt{\textbf{Prompt I}}
\begin{minted}[breaklines=true, bgcolor=promptColor1]{jinja}
Identify all named entities in the sentence provided according to the available entity types. Use `$$` as a separator between each pair of identified named entity types and corresponding content from the sentence. Only return the listed named entities without providing any additional commentary.

# Output Format
- List all the named entities found in the passage provided by the user. 
- Separate the paired named entities types and text using a `$$` symbol.
- Only return the entity list, without any prefix or explanation.
\end{minted}
\texttt{\textbf{Prompt II}}
\begin{minted}[breaklines=true, bgcolor=promptColor2]{jinja}
Identify and extract named entities from the provided sentence. Each identified entity pair (including entity type and content from the sentence) should be separated from their content using the "$$" delimiter.

# Steps
1. Analyze the sentence to identify named entities.
2. Extract each identified named entity and its content.
3. Concatenate the named entity type and its content with space as one pair.
4. Join all pairs of named entities using "$$" as a delimiter.
\end{minted}
\texttt{\textbf{Prompt III}}
\begin{minted}[breaklines=true, bgcolor=promptColor3]{jinja}
Extract named entities from the provided text and format the output by placing $$ between each entity type and its respective content. Ensure the output contains only the extracted entities and their labels, with no additional commentary or information.

# Steps
1. Analyze the provided text and identify named entities.
2. Categorize each identified entity by its correct type, careful to match the entity with the appropriate label.
3. Format the output by placing the entity type and its corresponding content, separated by $$.
\end{minted}
\texttt{\textbf{Prompt IV}}
\begin{minted}[breaklines=true, bgcolor=promptColor4]{jinja}
Identify named entities from the provided text. Format each entity and its content using $$ as a separator. 

# Steps
1. Parse the input text to identify all named entities. This includes proper nouns like names of people, places, organizations, dates, etc.
2. For each identified entity, extract the specific text corresponding to the entity.
3. Concatenate the name of the entity type and the associated text using space. 
4. Compile these formatted entries into a list with the $$ as a separator.

# Output Format
- A string joined by a " $$ " for each pair of the entity type and content, formatted as `EntityType EntityContent`.
\end{minted}
\texttt{\textbf{Prompt V}}
\begin{minted}[breaklines=true, bgcolor=promptColor5]{jinja}
Detect named entities in the supplied sentence. Use $$ as a separator between entities and their corresponding parts of the sentence. Limit the response strictly to the formatted list.

# Output Format
- Entities and their parts separated by $$
- Return a plain list with no additional context
- If no entities are present, return `$$`
\end{minted}
\texttt{\textbf{Suffix for Zero-shot and Few-shot}}
\begin{minted}[breaklines=true]{jinja}
# Named Entities Types to Identify
ACCOUNT_TYPE, ARTIST_NAME, BANK_NAME, BILL_TYPE, CALENDAR_EVENT, CITY_OR_PROVINCE, COUNTRY, CURRENCY, DATE, DISH_OR_FOOD, HOTEL_NAME, LANGUAGE_NAME, MEAL_PERIOD, MONEY, MUSIC_GENRE, NUMBER, PAYMENT_COMPANY, PERSONAL_NAME, PLACE_NAME, RESTAURANT_NAME, SHOPPING_ITEM, SONG_NAME, TIME

{% if shot_count == 0 -%}
{# Zero-shot Suffix #}
Please ensure that the entities match the listed types and that unstated entities should not be included in the response if no entities are found, return `$$` only.

# Format Example:
Sentence: John went to Paris and paid 100 dollars at an Awater restaurant.
Output: PERSONAL_NAME John $$ CITY_OR_PROVINCE Paris $$ MONEY 100 $$ RESTAURANT_NAME Awater
{% else %}
{# Few-shot Suffix #}
# Output Examples (Do not include in the response):
{% for example in examples -%}
Sentence: {{ example.text }}
Output: {{ example.slot }}

{% endfor %}
Based on the example, consider the following:
{% endif %}
Sentence: {{ text }}
Output: 
\end{minted}


\section{Instruction for Annotators}
\label{app:annotation-instruction}

This section provides a brief introduction to the annotation guide for the Slot Filling task.
Follow the instructions below to annotate the input text accordingly.

\begin{minted}[breaklines=true]{markdown}

A Slot Filling task is a natural language processing (NLP) task that involves extracting specific pieces of information (slots) from a given text. This task is commonly used in dialogue systems and information extraction applications where the goal is to identify and fill predefined categories or slots with relevant information from user inputs or text data.

### LANGUAGE_NAME
1. Spanish: A Romance language that originated in the Iberian Peninsula and is now the primary language of Spain and most Latin American countries.
2. Luganda: A Bantu language spoken primarily in Uganda, particularly by the Ganda people.
3. French: A Romance language spoken as a first language in France, parts of Belgium, and Switzerland, and in various communities worldwide.


### ACCOUNT_TYPE
1. Savings Account: A bank account that earns interest over time, typically used for long-term savings.
2. Checking Account: A bank account used for everyday transactions, such as deposits and withdrawals.
3. Student Account: A bank account designed for students, often with no monthly fees and special benefits.
Not to be confused with payment company. 
A credit card is NOT an account type.

### MONEY
1. $500: Five hundred dollars, often used to signify a substantial amount of money in various contexts.
2. 5 dollars: A small amount of money, typically used for minor purchases or expenses.
3. $1,000: One thousand dollars, indicating a significant sum, commonly used in transactions or savings.

### CURRENCY
1. Dollar: The currency of several countries, including the United States, Canada, and Australia.
2. Euro: The official currency of the Eurozone, used by 19 of the 27 European Union member states.
3. Yen: The official currency of Japan.

### CITY_NAME
1. London: The capital city of the United Kingdom, known for its historical landmarks and cultural diversity.
2. Kampala: The capital city of Uganda, known for its bustling markets and vibrant cultural scene.
3. New York: A major city in the United States, known for its skyscrapers and as a global financial and cultural center.
If you are not sure if a place is a City name (Town name) State/Province or Village name, please refer to a search engine for clarification.

### FOOD_ITEM
1. Sugar: A sweet substance commonly used in baking and cooking.
2. Orange: A citrus fruit known for its sweet and tangy flavor and high vitamin C content.
Not to be confused with Shopping item or Dish name.

### BANK_NAME
1. Ecobank: A pan-African banking conglomerate with operations in 36 African countries.
2. Wells Fargo: An American multinational financial services company headquartered in San Francisco, California.
3. HSBC: A British multinational banking and financial services organization with global operations.
When annotating Bank names, you do not need to include “bank” unless it is attached to the bank name, like seen above, with Ecobank.

### RESTAURANT_NAME
1. KFC: An American fast-food restaurant chain known for its fried chicken.
2. McDonald's: An American fast-food company famous for its hamburgers, fries, and other quick-serve meals.
3. Subway: An American fast-food franchise known for its submarine sandwiches (subs) and salads.

### DISH_NAME
1. Jollof Rice: A popular West African dish made with rice, tomatoes, onions, and various spices.
2. Paella: A Spanish rice dish originally from Valencia, featuring saffron, meat, seafood, and vegetables.
3. Sushi: A Japanese dish consisting of vinegared rice accompanied by various ingredients such as raw fish and vegetables.

### TIME
1. 2pm: A specific time in the afternoon.
2. Morning: The period from sunrise until noon.
3. Evening: The period of the day from the end of the afternoon to the beginning of night.
Anything that is less than one day should be annotated as TIME and not DATE, as seen in the above examples.

### TIMEZONE
1. Pacific Time (PT): A time zone covering parts of western Canada, the western United States, and western Mexico.
2. West Africa Time (WAT): A time zone used by countries in West Africa, one hour ahead of Coordinated Universal Time (UTC+1).
3. Eastern Standard Time (EST): A time zone covering parts of the eastern United States and parts of Canada, five hours behind Coordinated Universal Time (UTC-5).

### DATE
1. January: The first month of the year in the Gregorian calendar.
2. 2024: A specific year.
3. October: The tenth month of the year in the Gregorian calendar.
Anything that is more than one day must be annotated as DATE and not time, as seen above 

### BILL_TYPE
1. Internet Fees: Charges for the provision of internet services.
2. School Fees: Costs associated with attending an educational institution.
3. Electricity Bill: Charges for the consumption of electrical power.
4. Water Bill: 
You include “bill” as part of the annotation.

### PLUG_TYPE
1. Type A: A two-pronged plug commonly used in North America and Japan.
2. Type C: A two-pin plug used in Europe, South America, and Asia.
3. Type G: A three-pronged plug used in the United Kingdom and other countries.
Internet cable, extension cord are NOT  plug types. 

### COUNTRY
1. Germany: A country in Central Europe known for its rich history and economic strength.
2. Nigeria: A country in West Africa, known for its diverse cultures and large population.
3. Japan: An island nation in East Asia known for its technology and rich cultural heritage.

### PERSONAL_NAME
1. Dave: A common given name.
2. Maria: A common given name, often used in Spanish and Portuguese-speaking countries.
3. Akiko: A common Japanese given name.
4. Don’t annotate titles as personal names e.g Mr., Dr., Mrs. 
Mom, dad, aunt, sister is NOT a personal names

### MUSIC_GENRE
1. Fuji: A popular Nigerian musical genre that originated from the Yoruba people.
2. Gospel: A genre of Christian music.
3. Rock: A broad genre of popular music that originated as "rock and roll" in the United States in the late 1940s and early 1950s.
Old songs are not genres- Do not annotate them

### ARTIST_NAME
1. Fela: Refers to Fela Kuti, a Nigerian multi-instrumentalist and pioneer of Afrobeat music.
2. Beyoncé: An American singer, songwriter, and actress.
3. Mozart: Wolfgang Amadeus Mozart, an influential classical composer from Austria.

### HOTEL_NAME
1. Radisson: A global hotel chain known for its upscale accommodations and services.
2. Marriott: A worldwide hospitality company with a broad range of hotels and related services.
3. Hilton: A global brand of full-service hotels and resorts.
You can annotate Radisson Hotel as a whole.

### MEAL_PERIOD
1. Breakfast: The first meal of the day, typically eaten in the morning.
2. Lunch: A meal eaten around midday.
3. Dinner: The main meal of the day, usually eaten in the evening.

### PAYMENT_COMPANY
1. Paypal: An American company operating a worldwide online payments system.
2. Stripe: An Irish-American financial services and software as a service (SaaS) company.
3. Visa: A multinational financial services corporation known for its credit and debit cards.
Not to be confused with account type.

### CONTINENT
1. Africa: The second-largest and second-most-populous continent on Earth.
2. Europe: A continent located entirely in the Northern Hemisphere and mostly in the Eastern Hemisphere.
3. Asia: The largest and most populous continent, located primarily in the Eastern and Northern Hemispheres.

### AIRPORT_NAME
1. Bole Addis Ababa International Airport: The main international gateway to Addis Ababa, Ethiopia.
2. Heathrow Airport: A major international airport in London, United Kingdom.
3. John F. Kennedy International Airport: A major international airport in New York City, United States.

### SUPERMARKET
1. Shoprite: A leading food retailer in Africa with stores in several countries.
2. Walmart: A large multinational retail corporation operating a chain of hypermarkets.
3. Tesco: A British multinational groceries and general merchandise retailer.

### STATE/PROVINCE
1. Quebec Province: A province in eastern Canada, the largest in area and second-largest in population.
2. Ogun State: A state in southwestern Nigeria.
3. California: A state in the western United States, known for its diverse geography and large economy.

### NUMBER
1. 10: A numerical value, often used to denote quantity or ranking.
2. 20: A numerical value, commonly used to signify quantity or sequence.
3. Fifty-four: non-numeric should be annotated as a number.

### NATIONALITY
1. Nigerian: Pertaining to Nigeria or its people.
2. Kenyan: Pertaining to Kenya or its people.
3. American: Pertaining to the United States of America or its people.

### CALENDAR_EVENT
1. Football Match: A scheduled competitive game of football (soccer).
2. Concert: A live music performance.
3. Wedding: A ceremony where two people are united in marriage.
Christmas, Valentines day, birthdays, etc

### SHOPPING_ITEM
1. Shoe: A covering for the foot, typically made of leather, having a sturdy sole and not reaching above the ankle.
2. Shirt: A piece of clothing worn on the upper body, typically with sleeves and a collar.
3. Laptop: A portable personal computer with a screen and alphanumeric keyboard.
Not to be confused with Food items

### SONG_NAME
1. African Queen: A popular song by Nigerian artist 2Baba.
2. Thriller: A song by Michael Jackson from his album of the same name.
3. Shape of You: A song by Ed Sheeran.

### CAR_TYPE
1. BMW: A German multinational company that produces luxury vehicles and motorcycles.
2. Sedan: A passenger car in a three-box configuration with separate compartments for the engine, passenger, and cargo.
3.SUV: A sport utility vehicle, typically equipped with four-wheel drive for on- or off-road ability.
Ambulance, Fire truck are not  car types.

### PLACE
1.Tourist Attractions: Places of interest that draw visitors due to their cultural, historical, natural, or recreational significance. Examples include the Eiffel Tower in Paris, a global cultural icon of France, and the Grand Canyon in Arizona, known for its immense size and its intricate and colorful landscape.

2. Museums: Institutions that collect, preserve, and display objects of historical, cultural, artistic, or scientific importance. Examples include the Louvre Museum in Paris, which houses a vast collection of art, and the Smithsonian National Museum of Natural History in Washington, D.C., known for its exhibits on natural history and anthropology.

3. Mall:  A large indoor shopping complex featuring a variety of retail stores, restaurants, and entertainment facilities. Examples include the Mall of America in Minnesota, which is one of the largest malls in the United States, and the Dubai Mall in the UAE, known for its luxury shops and attractions like the Dubai Aquarium and Underwater Zoo.

4. Park:  A public area set aside for recreation and enjoyment, often featuring green spaces, playgrounds, and walking paths. Examples include Central Park in New York City, a vast urban park offering numerous recreational activities, and Hyde Park in London, known for its historical significance and open-air concerts.

With this entity, only annotate if entity is named explicitly, e,g Name of airport, museum or mall is not and nt just “mall”, “airport” etc

PS: Do not skip any annotations, if there is nothing to annotate, submit and go to the next one.
\end{minted}
\end{document}